\documentclass[journal]{IEEEtran}

\usepackage[T1]{fontenc}
\usepackage[utf8]{inputenc}
\usepackage{amsmath,amssymb,amsfonts}
\usepackage{graphicx}
\usepackage{booktabs}
\usepackage{array}
\usepackage{textcomp}
\usepackage{xcolor}
\usepackage{algorithm}
\usepackage{algpseudocode}
\usepackage{enumitem}
\usepackage[group-separator={,},group-minimum-digits=4]{siunitx}
\usepackage{url}
\usepackage[hidelinks]{hyperref}
\usepackage{orcidlink}

\begin{document}

\title{Lightweight Distillation of SAM~3 and DINOv3 for
Edge-Deployable Individual-Level Livestock Monitoring and
Longitudinal Visual Analytics}

\author{Haiyu~Yang\,\orcidlink{0009-0007-5125-4958}
        and~Miel~Hostens\,\orcidlink{0000-0001-5376-976X}%
\thanks{H. Yang and M. Hostens are with the College of Agriculture and
Life Sciences, Cornell University, Ithaca, NY 14853 USA.}%
\thanks{Corresponding author: Haiyu Yang (e-mail: hy625@cornell.edu).}%
\thanks{This work did not receive any specific grant from funding
agencies in the public, commercial, or not-for-profit sectors.}}

\markboth{IEEE Access}%
{Yang \MakeLowercase{\textit{et al.}}: Lightweight Distillation of SAM~3 and DINOv3 for Edge-Deployable Livestock Monitoring}

\maketitle

\begin{abstract}
Foundation-model pipelines have raised the accuracy ceiling of precision
livestock farming (PLF), but their GPU memory budgets exceed the envelope
of commodity edge accelerators, which has so far prevented their
deployment on hardware that can be installed in a barn. This paper
presents a knowledge-distillation and inference-compression method that
closes this gap for individual-level livestock monitoring. The
446\,M-parameter Perception Encoder (PE-ViT-L+) backbone of SAM~3 is
distilled into a 40.66\,M-parameter multi-scale student through three
contributions: a Feature Pyramid Network student encoder built on
TinyViT-21M-512; a four-term direction-then-scale distillation loss that
decouples directional alignment from scale calibration; and
backbone-substitution inference with sliding-window session pruning that
bounds streaming GPU-memory growth. We further establish empirically that
the pre-distilled DINOv3-ViT-S/16 variant (21.6\,M parameters) is a
sufficient per-individual embedder. On the Edinburgh Pig dataset, the
compressed pipeline reaches \SI{92.29}{\percent} MOTA and
\SI{96.15}{\percent} IDF1 relative to the SAM~3 teacher (losses of 1.68
and 0.84 percentage points), a 7.77-fold reduction in system-level
parameters, and a 3.01-fold reduction in peak VRAM
($\SI{19.52}{\giga\byte}\rightarrow\SI{6.49}{\giga\byte}$), while
attaining \SI{97.34}{\percent} top-1 accuracy and \SI{91.67}{\percent}
macro-F1 on nine-class pig-behaviour classification. The pipeline fits
inside an NVIDIA Jetson Orin~NX \SI{16}{\giga\byte} envelope with
\SI{4.9}{\giga\byte} of headroom, supporting a proposed --- but not yet
empirically validated --- on-device embedding-pool re-identification
mechanism whose per-individual footprint of approximately
\SI{94}{\mega\byte} per animal per year yields a longitudinal visual
record amenable to retrospective association with disease, lameness,
reproductive, and growth outcomes.
\end{abstract}


\IEEEpeerreviewmaketitle

\section{Introduction}

\subsection{Individual-level monitoring as the new target of precision livestock farming}

Precision livestock farming (PLF) has been transforming from a herd-level
data-collection discipline into an individual-level decision-support
discipline. Reviews over the past decade
\cite{berckmans2014precision,neethirajan2021digital,papakonstantinou2024precision,rocchi2025precision}
converge on a clear trend: welfare assessments, disease detection, and
productivity forecasts all improve when behaviour is attributed to specific
animals rather than aggregated across a group. At the same time,
computer-vision pipelines have become the dominant non-invasive channel for
collecting such individual-level data, because they avoid the stress,
battery-life, and maintenance costs of wearable sensors while still
producing per-individual signals \cite{neethirajan2021digital,bello2024computer}.
Group-housed species, where individuals mix freely and share space, feed,
and enrichment, remain the hardest setting for this transition. Pigs in
particular occlude one another, adopt near-identical postures during rest,
and are difficult to distinguish by coat pattern, so maintaining consistent
per-individual identity over hours of video has been the limiting factor in
operationalising the technology
\cite{bergamini2021extracting,yang2025computer}.

\subsection{Foundation-model pipelines have raised the accuracy ceiling}

Recent foundation-model developments have pushed per-individual tracking
and behaviour recognition substantially closer to production quality.
Open-vocabulary detectors such as OWLv2 \cite{minderer2023scaling} detect
pigs reliably in overhead footage without task-specific annotation;
promptable video-segmentation models such as SAM~2 \cite{ravi2024sam2}
and more recently SAM~3 \cite{carion2025sam3} track individuals across
minutes of video from a single bounding-box prompt; and self-supervised
visual backbones such as DINOv2 \cite{oquab2024dinov2} and DINOv3
\cite{simeoni2025dinov3} produce per-crop embeddings whose quality on
behaviour-classification tasks rivals or exceeds task-specific supervised
features. Our prior work \cite{yang2025computer} demonstrated that
assembling such foundation models into a modular pipeline ---
open-vocabulary detection, motion-aware video segmentation, self-supervised
embedding, and a light temporal classifier --- reaches \SI{94.2}{\percent}
behaviour-classification accuracy on the Edinburgh Pig Behaviour Video
Dataset \cite{bergamini2021extracting}, a 21.2-percentage-point
improvement over the previous state of the art on the same benchmark.
That pipeline serves as the reference system from which the present work
departs.

\subsection{The edge-deployment gap}

The accuracy gains of foundation-model pipelines come with a compute gap
that has so far prevented their deployment on the hardware that can
realistically be installed in a barn. Assembling a SAM~3 tracker with a
DINOv3-ViT-7B embedder co-resident on a single GPU produces a memory
footprint that exceeds the \SI{16}{\giga\byte} envelope of the NVIDIA
Jetson Orin~NX \cite{nvidia2023orin}. Our SAM~3 inference reached a
measured peak of \SI{19.52}{\giga\byte} on a single NVIDIA A10
(\S\ref{sec:efficiency}); the published DINOv3-ViT-7B model adds another
\SI{12.51}{\giga\byte} at fp16 or \SI{25.02}{\giga\byte} at fp32 in
weights alone (6716\,M parameters $\times$ 2 or 4~B per
parameter), before activations and tracking-session memory are considered.
Continuous inference must currently be routed to a cloud GPU instance or
to a workstation in a farm office, which introduces connectivity, latency,
and recurring-cost problems poorly suited to rural deployments
\cite{rocchi2025precision}. Even on a cloud A10 GPU, SAM~3's session-based
inference at native resolution produces a tracker session that grows
monotonically in memory: in our measurements, the session consumed roughly
\SI{5.6}{\mega\byte} per frame per tracked object. 
Edge-ready pipelines
reported for grassland cattle (MASM-YOLO on Jetson Orin~NX at \SI{36}{\hertz})~\cite{wei2025lightweight} and for crop-harvesting robots
\cite{kim2025real} achieve on-device real-time performance by starting
from small, specialised detectors such as YOLOX or YOLOv12 rather than
foundation models, which sidesteps the compute gap but also forfeits the
accuracy ceiling that foundation-model pipelines have established.

\subsection{Compression for SAM-family models: relevant prior art}
\label{sec:priorart}

A body of work on compressing the Segment Anything Model (SAM;
\cite{kirillov2023segment}) family has emerged in response to
precisely this compute gap. We organise that work into three generations
that match the three SAM releases.

\paragraph{SAM~1 distillations.} MobileSAM \cite{zhang2023faster}
pioneered decoupled knowledge distillation from the SAM ViT-H image
encoder (632~M parameters) into a TinyViT-based student (9.8~M parameters)
and reached 38.7 COCO zero-shot mAP. EfficientSAM \cite{xiong2023efficientsam}
used masked-image pretraining as a distillation strategy and reached 44.4
mAP with 25.3~M encoder parameters. EdgeSAM \cite{zhou2024edgesam}
distilled into a purely CNN-based encoder with prompt-in-the-loop
supervision, achieving a 37-fold speedup on desktop GPUs.
EfficientViT-SAM \cite{zhang2024efficientvitsam} replaced the encoder
with the EfficientViT architecture and achieved up to a 48.9-fold
throughput gain on TensorRT while matching SAM-ViT-H zero-shot accuracy.
All four target single-image segmentation under natural-image prompts and
validate on COCO, LVIS, and SA-1B.

\paragraph{SAM~2 / SAM~3 distillations.} The video-tracking generation of
SAM introduces a second computational bottleneck beyond the image
encoder --- the dense memory-attention module that maintains identity
across frames --- and the SAM-1-era recipes do not address it. Two recent
efforts close that gap. EdgeTAM and EfficientTAM (referenced in
\cite{zeng2025efficientsam3}) compress SAM~2's encoder and memory
modules for streaming use. EfficientSAM3 \cite{zeng2025efficientsam3},
released contemporaneously with SAM~3 itself, introduces a three-stage
Progressive Hierarchical Distillation recipe: encoder distillation on
SA-1B, Perceiver-based temporal-memory distillation on SA-V, and
end-to-end fine-tuning on SAM~3's PCS data. They release a model zoo of
nine students using RepViT, TinyViT, and EfficientViT backbones,
benchmarked on SA-Co, SA-V, COCO, and standard VOS datasets.

\paragraph{What the present work targets.} The two SAM-3-era efforts above
both target general-purpose promptable concept segmentation, validated
against natural-image and video-object-segmentation benchmarks. None of
the existing distillations --- across either generation --- reports the
two metrics that matter operationally for per-animal welfare monitoring
on overhead livestock footage: identity preservation under occlusion
across minutes of video (MOTA, IDF1), and downstream behaviour-classification
quality of the compressed pipeline. The Edinburgh Pig benchmark
exercises failure modes that natural-image benchmarks do not: heavy
intra-class similarity, frequent partial occlusion, and near-identical
postures during rest. The present paper therefore positions itself
not as a competing general-purpose distillation, but as the first
SAM-distillation effort whose target deliverable is per-animal welfare
analytics on group-housed livestock, and the first published distilled
SAM~3 checkpoint specialised for pig video.

\subsection{Compression versus replacement: when to distill and when to substitute}

A second observation motivated our two-track approach. The DINOv3 family
\cite{simeoni2025dinov3} was released with a 7~B-parameter ViT-7B
teacher together with several publicly available pre-distilled smaller
variants, of which the smallest is ViT-S/16 (21.6\,M
parameters), produced by Meta through a full distillation pass on the
same LVD-1689M pretraining corpus. Re-distilling such an
already-distilled model within our own pipeline would at best recover
the quality of the existing ViT-S checkpoint and at worst degrade it.
The correct question for DINOv3 is therefore empirical rather than
architectural: does the pre-distilled ViT-S variant retain enough
representational capacity for our downstream behaviour-classification
task? The analogous question does not have a ready-made answer for SAM~3,
because no publicly available distilled variant of SAM~3 specialised for
overhead livestock footage existed at the time of our work. The
compression track for SAM~3 and the adoption track for DINOv3-ViT-S
therefore require different treatments within the same paper.

\subsection{Contributions}
\label{sec:contributions}

\begin{description}[leftmargin=1em,labelindent=0em,style=unboxed]
  \item[\normalfont(C1) Distillation of SAM~3's vision backbone into a
    lightweight multi-scale student encoder with minimal tracking
    degradation.] We compress SAM~3's 446\,M-parameter Perception
    Encoder (PE-ViT-L+) image backbone into a 40.66\,M-parameter
    student --- a 10.98-fold reduction at the backbone and a 7.77-fold
    reduction at the assembled-system level --- with only
    1.68-percentage-point loss in MOTA and 0.84-percentage-point loss in
    IDF1 on held-out overhead pig-video clips. This result rests on
    three mechanisms introduced in Section~\ref{sec:methods}: a
    multi-scale student encoder (StudentEncoderTinyViT) that fuses three
    stages of a TinyViT-21M-512 backbone \cite{wu2022tinyvit} rather
    than projecting a single stage; a four-term direction-then-scale
    distillation loss that decouples directional alignment from scale
    calibration; and backbone-substitution inference together with a
    sliding-window session-pruning routine that bounds streaming GPU
    memory to the most recent eight non-conditioning frame outputs per
    object.

  \item[\normalfont(C2) Empirical validation of DINOv3-ViT-S (21M) as a
    behaviour-discriminative embedder.] Adopting the pre-distilled
    ViT-S/16 variant --- the smallest publicly released DINOv3 model,
    with 21.6\,M parameters --- we train only a downstream
    temporal classifier and report \SI{97.34}{\percent} top-1 accuracy
    with \SI{91.67}{\percent} macro-F1 on nine pig behaviours across
    \num{4292} test windows. To our knowledge this is the first reported
    empirical quantification of DINOv3-ViT-S on a precision-livestock-farming
    task, establishing the smallest DINOv3 variant as a viable embedder
    for on-device PLF pipelines.

  \item[\normalfont(C3) An embedding-pool mechanism for longitudinal
    per-individual visual analytics.] We design and analytically size an
    embedding-pool re-identification loop that maintains stable identity
    for each animal across sessions --- day to day and week to week, not
    only within a single video clip --- by using DINOv3-ViT-S embeddings
    as a long-horizon identity signal. At an hourly sampling cadence the
    mechanism produces, for each animal, a time-indexed embedding history
    tagged with timestamp and behaviour context, occupying approximately
    \SI{94}{\mega\byte} per individual per year and approximately
    \SI{19}{\giga\byte} per year for a 200-animal barn. We present this
    mechanism as a substrate that future work can build on, and scope
    both the re-identification loop's empirical identity-switch
    performance and any specific downstream analytic as follow-up work
    (\S\ref{sec:followup}). The mechanism is species-agnostic and
    extends straightforwardly from pigs to dairy cattle, beef cattle,
    sheep, and other commercially important group-housed species.

  \item[\normalfont(C4) Integrated edge-device feasibility audit.]
    Combining the SAM~3 distillation result with the DINOv3-ViT-S
    adoption, the compressed pipeline's peak VRAM is
    \SI{6.49}{\giga\byte} on an NVIDIA A10. Combined with published
    memory budgets for OWLv2, the BiLSTM classifier, and the CUDA
    runtime, the full pipeline fits inside the \SI{16}{\giga\byte}
    envelope of a Jetson Orin~NX with \SI{4.9}{\giga\byte} of
    headroom --- sufficient to operate the embedding-pool mechanism of
    C3 on-device without external compute.
\end{description}

\subsection{Paper organisation}

Section~\ref{sec:methods} describes the dataset, the teacher pipeline,
the student architecture, the DINOv3 adoption, the proposed
embedding-pool re-identification mechanism, and the evaluation and
computational protocols. Section~\ref{sec:results} reports quantitative
results, organised into SAM~3 distillation (\S\ref{sec:sam3results}),
DINOv3-ViT-S adoption (\S\ref{sec:dinov3results}), and an analytic
edge-device feasibility audit (\S\ref{sec:edgefeasibility}).
Section~\ref{sec:discussion} discusses the results, addresses the limits
of our per-video analysis and the scope of the deferred re-identification
validation, and outlines the principal directions for follow-up work.
Section~\ref{sec:conclusions} concludes. Appendices A--C provide the
session-pruning pseudocode, the re-identification pseudocode, and a
hyperparameter reference.

\section{Material and methods}
\label{sec:methods}

We describe here the dataset, our prior teacher pipeline, the
student-encoder architecture and distillation protocol for SAM~3, the
adoption of DINOv3-ViT-S~(21M) as the per-individual embedder, the
proposed embedding-pool re-identification mechanism, the evaluation
metrics, and the hardware and software environment used for all training
and benchmarking. The section is organised so that
\S\ref{sec:teacherpipeline} establishes the reference system,
\S\ref{sec:sam3distill}--\S\ref{sec:dinov3adopt} describe the two
compression-and-adoption targets in detail, \S\ref{sec:reid} introduces
the forward-looking re-identification loop that the compressed pipeline
enables, and \S\ref{sec:metrics}--\S\ref{sec:hardware} fix the
measurement and computational protocols used throughout
Section~\ref{sec:results}.

\subsection{Dataset and task}
\label{sec:dataset}

All experiments were conducted on the Edinburgh Pig Behaviour Video
Dataset \cite{bergamini2021extracting}, publicly available at
\url{https://homepages.inf.ed.ac.uk/rbf/PIGDATA/}. The dataset was
collected over six weeks in late 2019 in a single
\SI{5.8}{\meter}~$\times$~\SI{1.9}{\meter} pen at Scotland's Rural
College research pig unit, with eight growing pigs monitored by an
overhead Intel RealSense D435i camera mounted \SI{2.5}{\meter} above
the floor. The present work used the annotated subset comprising nine
video clips of 600 frames each, at a native resolution compatible with
$1280 \times 720$ capture, with per-frame bounding boxes and consistent
identity labels for all eight individuals in each clip. In aggregate,
the nine clips yielded \num{5400} annotated frames and \num{43200}
individual-pig bounding-box annotations.

We used the dataset in two distinct ways, reflecting the two targets of
this study. For SAM~3 distillation, we partitioned the nine clips into
four training clips (\texttt{2019\_11\_05\_000002},
\texttt{2019\_11\_11\_000028}, \texttt{2019\_11\_11\_000036},
\texttt{2019\_11\_22\_000010}), one validation clip
(\texttt{2019\_11\_28\_000113}), and four held-out test clips
(\texttt{2019\_12\_02\_000005}, \texttt{2019\_12\_02\_000208},
\texttt{2019\_12\_10\_000060}, \texttt{2019\_12\_10\_000078}). The
split was fixed at the clip level rather than the frame level to
prevent information leakage through temporal adjacency between training
and evaluation frames. Distillation targets were extracted once per
training and validation frame, as described in
\S\ref{sec:targetrep}, and reused across every training epoch.

For the DINOv3 behaviour-classification task, we used per-pig cropped
frames extracted from all nine clips through our prior pipeline
\cite{yang2025computer}. Crops were assigned behaviour labels drawn
from a nine-class label set: \emph{standing, lying, eat, drink, sitting,
sleep, run, playwithtoy,} and \emph{nose-to-nose}. The resulting
dataset contained approximately \num{43000} labelled crops with a
highly non-uniform class distribution --- in particular, \emph{sleep}
dominated at roughly \SI{53}{\percent} of windows, while \emph{run}
and \emph{playwithtoy} were rare at less than \SI{0.5}{\percent} of
windows each. This imbalance is a property of the underlying behaviour
distribution in group-housed pigs rather than an artefact of the
labelling protocol, and we handled it explicitly through class-weighted
training (\S\ref{sec:bilstm}) rather than through resampling.

\subsection{Teacher pipeline}
\label{sec:teacherpipeline}

The starting point of the present work is the modular vision pipeline
we introduced in our earlier publication \cite{yang2025computer},
which operates in four sequential stages: open-vocabulary detection of
pigs using OWLv2 \cite{minderer2023scaling}; per-individual
segmentation and temporal tracking using a video-segment-and-track
model; per-crop feature extraction using a self-supervised vision
transformer; and behaviour classification using a Long Short-Term
Memory (LSTM) network operating over short sequences of per-individual
embeddings. On the Edinburgh Pig dataset, that pipeline reached
\SI{94.2}{\percent} behaviour-classification accuracy at the sequence
level, representing a 21.2-percentage-point improvement over the
previous state of the art on the same benchmark.

For the present paper, we made two generational upgrades to the tracker
and embedder before attempting compression. First, we replaced the
SAM~2-era tracker used in our prior pipeline \cite{ravi2024sam2} with
SAM~3 \cite{carion2025sam3}, released by Meta in late 2025. SAM~3
introduces promptable concept segmentation and a unified
image-and-video architecture consisting of an image-level detector and
a memory-based video tracker that share a single Perception Encoder
(PE) vision backbone, delivering substantially more accurate
prompt-driven video segmentation and tracking than its predecessor.
Second, we replaced the DINOv2-ViT backbone \cite{oquab2024dinov2}
with DINOv3 \cite{simeoni2025dinov3}, which provides higher-quality
dense features through an updated training recipe and a 1.689-billion-image
pretraining set (LVD-1689M). For the DINOv3 stage we adopted the
pre-distilled DINOv3-ViT-S/16 directly (denoted DINOv3-ViT-S~(21M)
hereafter); we did not instantiate the 6716\,M-parameter
ViT-7B variant in the pipeline, because the pre-distilled smaller
variants are released by Meta together with the teacher and
re-distilling within our own pipeline would yield no benefit. The
resulting reference pipeline ---
$\textrm{OWLv2} \rightarrow \textrm{SAM~3} \rightarrow
\textrm{DINOv3-ViT-S~(21M)} \rightarrow \textrm{BiLSTM}$ ---
preserves the modular structure of our prior work while being more
accurate at each stage. The two heaviest modules of the reference
system are SAM~3 (measured peak \SI{19.52}{\giga\byte} on A10;
\S\ref{sec:efficiency}) and the SAM~3 streaming session that grows
linearly with frame count; the remainder of this section describes how
we brought SAM~3 within the \SI{16}{\giga\byte} Jetson Orin~NX envelope
while preserving the pipeline's accuracy.

\subsection{SAM 3 backbone distillation}
\label{sec:sam3distill}

\subsubsection{Target representation}
\label{sec:targetrep}

SAM~3's vision module consists of a Perception Encoder (PE-ViT-L+)
backbone (approximately 446\,M parameters) followed by a
four-level Feature Pyramid Network (FPN) neck, a prompt-conditioned
mask decoder, and a cross-frame memory-attention module. A natural
first question in designing a distillation target is where along this
path to match the student to the teacher. Matching at the decoder
logits preserves the full teacher computation graph but forces the
student to also absorb the neck, decoder, and memory modules --- a much
harder learning problem that would sacrifice most of the compression
benefit. Matching at the raw backbone output, before the neck, is the
opposite extreme: it localises the capacity reduction to the encoder
and leaves the rest of SAM~3's pretrained machinery untouched at
inference. We took the second approach.

Our distillation target is the PE-ViT-L+ backbone output at the end of
the last transformer block, reshaped to spatial form as a tensor of
shape $[1, 1024, 72, 72]$ at the native $1024 \times 1024$ input
resolution. Each training and validation frame was passed once through
the teacher, the backbone output was captured through a PyTorch
forward hook, cast to float16 to halve the storage cost, and written
to disk as a \texttt{.pt} tensor. This yielded \num{2400} training
targets and 600 validation targets, each occupying approximately
\SI{10}{\mega\byte} on disk.

\subsubsection{Student architecture (StudentEncoderTinyViT)}
\label{sec:studentarch}

The main architectural contribution of this paper is a compact student
encoder, which we denote StudentEncoderTinyViT, built on top of the
TinyViT-21M-512 backbone \cite{wu2022tinyvit} available through the
\texttt{timm} library \cite{wightman2019timm}. TinyViT was originally
distilled from a large image-classification teacher at $512 \times 512$
resolution and combines convolutional stem and stage-one layers with
transformer stages two through four; this hybrid structure provides
both the local inductive bias needed for fine pig silhouettes and the
global context needed for group-housed scenes.

We used TinyViT in \texttt{features\_only} mode with
\texttt{out\_indices = (1, 2, 3)}, which exposes the outputs of stages
2, 3, and 4 at channel dimensions 192, 384, and 576, respectively.
Each stage output was passed through a lateral $1 \times 1$ convolution
followed by GroupNorm-32 \cite{wu2018groupnorm} and a GELU
nonlinearity \cite{hendrycks2016gaussian}, projecting it to a common
channel dimension of 256. The three projected stage maps were spatially
resampled to a common $72 \times 72$ grid --- matching the teacher
output's spatial resolution --- and concatenated along the channel
axis, yielding a 768-channel multi-scale fused representation. A
$1 \times 1$ convolution projected this fused map to the teacher's
1024-channel dimension, followed by a residual spatial-refinement block
consisting of two $3 \times 3$ convolutions with GroupNorm-32 and
GELU. The final output was modulated by a learnable per-channel affine
pair (\texttt{output\_scale}, \texttt{output\_bias}) initialised to
$(1, 0)$, which allowed the network to calibrate the global first and
second moments of its output without routing this calibration through
the convolutional stack. The full student output is a tensor of shape
$[B, 1024, 72, 72]$, exactly matching the teacher target representation
defined in \S\ref{sec:targetrep}. This design introduces approximately
40.66\,M parameters in total, of which the TinyViT backbone
accounts for 21.6\,M, the lateral-and-fusion projections for
approximately 7\,M, the channel projection for
0.8\,M, the spatial-refinement block for 9.4\,M, and
the affine output for \num{2048} parameters.

\subsubsection{Architectural design decisions}

Three design choices in \S\ref{sec:studentarch} deserve explicit
justification because they materially affect the quality of the
distilled representation.

First, we replaced every BatchNorm layer in the student --- including
the BatchNorm layers embedded deep inside the TinyViT backbone ---
with GroupNorm. BatchNorm \cite{ioffe2015batch} computes its
normalisation statistics over the batch dimension at training time and
uses stored running estimates at evaluation time; this creates a
train--eval statistical mismatch that is ordinarily benign for large
batches but becomes non-trivial at our training batch size of four
frames. More importantly, because our distillation targets were
themselves computed in the teacher's evaluation mode, any train--eval
mismatch on the student side produced a silent, systematic bias
between the input distribution seen during target matching and the
input distribution seen during subsequent inference. Replacing every
BatchNorm layer with GroupNorm \cite{wu2018groupnorm} using
$\min(32, C)$ groups eliminated this bias entirely, producing
bit-identical outputs between training and evaluation modes; we
verified the absence of any residual mismatch by checking that the
maximum element-wise absolute difference on a random input tensor was
below $10^{-5}$.

Second, we fused features from three stages rather than projecting
from a single stage. Pigs in the Edinburgh dataset are large objects
in the frame, but their fine silhouettes at the animal boundary ---
critical for SAM~3's mask prediction --- carry high-frequency content
that is poorly represented in any single stage of a hybrid backbone.
Lateral FPN-style fusion in the manner of Lin \textit{et al.}~\cite{lin2017feature}
combines the local detail of shallower stages with the semantic
richness of deeper stages at the target resolution.

Third, we zero-initialised the second GroupNorm layer of the residual
spatial-refinement block. This initialisation makes the residual
branch a no-op at the start of training, so the network first learns
to match teacher direction through the lateral-fusion path alone, and
the refinement block subsequently learns only the residual component
needed to close the remaining gap. The scheme, borrowed from the
fixup and zero-init residual-initialisation literature
\cite{zhang2019fixup,goyal2017accurate,bachlechner2021rezero},
accelerated convergence substantially in pilot experiments.

\subsubsection{Loss function}
\label{sec:lossfn}

The distillation objective combines four terms arranged so that
directional alignment with the teacher is learned first and scale
matching is learned second:
\begin{equation}
\label{eq:loss}
\begin{aligned}
\mathcal{L} ={} & 1.0 \cdot \left\| \frac{F_s}{\|F_s\|_2} - \frac{F_t}{\|F_t\|_2} \right\|_2^2 \\
& {}+ 0.5 \cdot \big( 1 - \cos(F_s, F_t) \big) \\
& {}+ 0.3 \cdot \big( \|\sigma_s - \sigma_t\|_2^2 + \|\mu_s - \mu_t\|_2^2 \big) \\
& {}+ 0.1 \cdot \| F_s - F_t \|_2^2 .
\end{aligned}
\end{equation}
Here $F_s \in \mathbb{R}^{B \times 1024 \times 72 \times 72}$ denotes
the student output, $F_t$ the teacher target of the same shape, and
$\mu_c$, $\sigma_c$ the per-channel mean and standard deviation over
the spatial axes. The first term is the mean-squared error between
$L_2$-normalised feature tensors and measures directional agreement at
the level of whole feature maps. The second term is a patch-wise
cosine-embedding loss; we reshaped both tensors to
$(B \cdot 72 \cdot 72) \times 1024$ before computing it, so that the
similarity constraint was applied at the level of each spatial location
independently. The third term encourages the student's per-channel
first and second moments to match the teacher's. The fourth term is
the raw un-normalised mean-squared error, retained at a low weight as
an auxiliary signal.

The specific weights ($1.0$ / $0.5$ / $0.3$ / $0.1$) were chosen
empirically on the basis of pilot training runs that exhibited a
collapsed-variance failure mode at higher relative weight on the
moment-matching term: when the scale-matching term was given equal
weight to the directional terms, the network tended to converge to an
ill-conditioned solution that matched the teacher's variance without
aligning with its direction. The weights reported here are the
configuration used to produce the results in Section~\ref{sec:results};
a systematic sweep over these weights is left to follow-up work.

\subsubsection{Optimisation protocol}

The student was trained with AdamW \cite{loshchilov2019decoupled}
using per-group learning rates: $1 \times 10^{-4}$ for the TinyViT
backbone, and $3 \times 10^{-4}$ for every other trainable parameter
group (the three lateral projections, the channel projection, the
spatial-refinement block, and the affine output). Weight decay was
set to $10^{-4}$ everywhere except on the affine parameters, which
received no weight decay to preserve their interpretability as scale
and bias. We used a cosine-annealing warm-restart schedule
\cite{loshchilov2017sgdr} with initial period $T_0 = 15$ epochs,
period multiplier $T_{\mathrm{mult}} = 2$, and minimum learning rate
$\eta_{\min} = 10^{-6}$. Gradient norms were clipped to 1.0 before the
optimiser step. Mixed-precision training was enabled using bfloat16
autocast \cite{micikevicius2018mixed}, which halved the memory
footprint of activations while preserving numerical stability on our
NVIDIA A10 GPU.

An exponential moving average (EMA) of the student parameters was
maintained in parallel with the online weights using decay 0.9995,
following the Polyak-averaging convention \cite{polyak1992acceleration}.
Validation loss was computed both on the online weights and on the
EMA shadow, and all reported student results in Section~\ref{sec:results}
are based on the EMA weights; in our runs the EMA consistently
outperformed the online weights by 1--3\% on validation cosine
similarity in late-training epochs. Early stopping was applied on the
EMA validation loss with patience 15 epochs, and the maximum number of
training epochs was 100.

Training used a batch size of four at the native $1024 \times 1024$
resolution. Data augmentation comprised horizontal flipping of the
image paired with a synchronised horizontal flip of the teacher target
tensor (applied with probability 0.5), and colour jitter with
brightness 0.2, contrast 0.2, saturation 0.1, and hue 0.02 applied to
the image only. We did not apply vertical flipping, rotation, or
crop-based augmentations, because SAM~3's teacher features encode
overhead-camera viewpoint assumptions that would be broken by any
reorientation of the scene. Best EMA validation loss was reached at
epoch 75 of a 100-epoch budget, and the checkpoint at epoch 75 was
used for all downstream evaluation.

\subsubsection{Inference-time integration (backbone substitution)}
\label{sec:backbonesub}

A convenient consequence of distilling at the raw backbone output
(\S\ref{sec:targetrep}) is that the trained student can be inserted
into SAM~3 by module substitution, without any modification to the
downstream neck, mask decoder, memory-attention, or memory-encoder
components. At inference, we replaced
\texttt{teacher\_model.vision\_encoder.backbone} with our trained
StudentEncoderTinyViT, wrapped in a small adapter class
(\texttt{DummyBackbone}) that presented the student's output tensor
$[B, 1024, 72, 72]$ in the $[B, N, C]$ contract expected by SAM~3's
\texttt{vision\_encoder.forward}. Specifically, the adapter reshaped
the spatial student output to $[B, 72{\times}72, 1024]$ and packed it
into a \texttt{BaseModelOutput} with the student tensor as
\texttt{last\_hidden\_state}. The remainder of
\texttt{vision\_encoder.forward} --- the position-encoding application
and the neck --- then processed the student features exactly as it
would have processed the teacher features.

The original teacher backbone was moved to CPU rather than deleted,
so that teacher and student could be benchmarked against each other
on the same hardware with a single module-swap operation. This
allowed every per-frame latency and peak-VRAM measurement reported in
Section~\ref{sec:results} to be made on identical input frames under
identical stateful conditions.

\subsubsection{Memory management during streaming}
\label{sec:memory}

SAM~3 performs video-object tracking through an inference-session
object that accumulates per-object mask outputs across frames; the
memory-attention module consumes these accumulated outputs to maintain
identity across time. In its default configuration, the session
retains every frame's output for every tracked object for the entire
video, producing monotonically growing GPU memory consumption. We
characterised this growth in a pilot measurement on a 60-frame prefix
of a test clip with eight tracked objects and observed linear growth
of approximately \SI{5.6}{\mega\byte} per frame per object after the
initial conditioning frame. Extrapolated to a \SI{30}{\hertz} live
stream, the session would consume the full \SI{16}{\giga\byte} of a
Jetson Orin~NX \cite{nvidia2023orin} in roughly 12 minutes of
continuous operation --- before any other module was loaded.

We implemented a sliding-window pruning routine, which we call
\texttt{prune\_inference\_session}, that retained only the most recent
eight non-conditioning frame outputs per object, removing older
entries from the per-object \texttt{non\_cond\_frame\_outputs} cache
at every 25-frame interval. The choice of eight follows SAM~3's
internal \texttt{num\_maskmem} constant, which bounds the number of
past frames the memory-attention module actually reads from; any
cached output beyond this window contributes no information to future
predictions. The pruning mechanism is a property of the inference
loop and is independent of whether the teacher or student backbone is
in use; both configurations benefit from it. Full pseudocode is
provided in Appendix~\ref{app:pruning}.

\subsection{DINOv3-ViT-S (21M) adoption}
\label{sec:dinov3adopt}

\subsubsection{Rationale}

The DINOv3 family \cite{simeoni2025dinov3} includes a
6.7-billion-parameter ViT-7B teacher with 4096-dimensional patch and
global embeddings, together with several smaller variants pre-distilled
from that teacher on the same LVD-1689M pretraining corpus. The
smallest publicly released variant is DINOv3-ViT-S/16, with
21.6\,M parameters and 384-dimensional embeddings. Because
this model was already distilled by Meta against the ViT-7B teacher,
an additional distillation pass in our pipeline would yield no
benefit. Instead, the question we addressed is whether the
pre-distilled ViT-S model is of sufficient representational quality
for the specific downstream task in our pipeline --- nine-class pig
behaviour classification from per-individual cropped image streams ---
and we answered it empirically by training a downstream temporal
classifier on ViT-S embeddings extracted from the Edinburgh Pig
dataset.

\subsubsection{Embedding extraction}
\label{sec:extract}

Per-individual crops were produced upstream of DINOv3 by the
OWLv2--SAM~3--bounding-box-crop chain described in
\S\ref{sec:teacherpipeline}, at one crop per (frame, individual) pair.
We processed each crop at the native DINOv3-ViT-S/16 input resolution
of $224 \times 224$ using the model's provided
\texttt{AutoImageProcessor} \cite{wolf2020transformers} for resizing
and normalisation. Inference was performed in float16 on a single GPU
at batch size 256; at this batch size DINOv3-ViT-S occupies
approximately \SI{1.5}{\giga\byte} of GPU memory. The pooled output
(the CLS-token-equivalent representation emitted at the end of the
network) of shape $[384]$ was stored as one \texttt{.pt} tensor per
crop, yielding files of approximately \SI{1.5}{\kilo\byte} each in
float32 and \SI{0.8}{\kilo\byte} each in float16. I/O-parallel
extraction using 36 worker threads for image loading and saving
allowed us to saturate the GPU throughout the extraction phase. The
full extraction of all \num{43000} crops completed in a single pass at
sustained throughput; the detailed benchmark is reported in
Section~\ref{sec:results}.

\subsubsection{Behaviour classifier (BiLSTM)}
\label{sec:bilstm}

For behaviour classification we trained a lightweight bidirectional
LSTM classifier
\cite{hochreiter1997long,schuster1997bidirectional} over short
sequences of consecutive per-individual embeddings. We used a
sequence length of three frames with stride one, sampled along the
per-individual embedding stream; each window was labelled with the
majority behaviour label among its three frames. We chose the BiLSTM
over heavier sequence models because the short sequence length and
small embedding dimension make attention-based aggregation
unnecessary, and a single-layer BiLSTM with hidden size 128 was
sufficient to saturate validation accuracy in pilot experiments.

The classifier architecture was as follows: a single-layer
bidirectional LSTM with input dimension 384 and hidden dimension 128
(approximately 560\,k parameters), followed by
$\textrm{Linear}(256, 128) \rightarrow \textrm{ReLU} \rightarrow
\textrm{Dropout}(0.3) \rightarrow \textrm{Linear}(128, 9)$, with
dropout regularisation \cite{srivastava2014dropout}. The last
timestep's bidirectional hidden state was used as the sequence
representation. Training used Adam \cite{kingma2015adam} with
learning rate $10^{-3}$, weight decay $10^{-5}$, batch size 128,
class-weighted cross-entropy using inverse-frequency weights on the
training label distribution, and early stopping with patience 8
epochs. We split the window-level dataset stratified on label into
$70/15/15$ training, validation, and test partitions. Because windows
overlap with stride one, some leakage of information across adjacent
windows is possible at the frame level; we mitigated this by splitting
at the window level after windowing rather than at the frame level
before windowing, which preserves the stratified label balance.

\subsection{Embedding-pool re-identification}
\label{sec:reid}

Even at Multi-Object Tracking Accuracy (MOTA) scores above
\SI{92}{\percent}, SAM~3 produces identity switches at a rate of
approximately one per 500 frames per animal under the conditions of
the Edinburgh dataset. Over a single 24-hour deployment at
\SI{5}{\hertz}, this accumulates into materially wrong per-individual
daily-budget statistics. To address this failure mode without
introducing supervised re-identification training, we propose an
embedding-pool re-identification mechanism that treats DINOv3
embeddings as a long-horizon identity signal complementary to SAM~3's
short-horizon motion-based tracking. The mechanism is proposed and
analytically sized in the present paper; empirical validation on
labelled long-duration video is scoped as follow-up work
(\S\ref{sec:followup}).

Formally, for each tracked identity $i$ we maintain an embedding bank
$E_i = \{e_i(t_0), e_i(t_1), \ldots\}$ consisting of
exponentially-moving-average (EMA) summaries of the DINOv3 embeddings
observed for identity $i$ at a configurable sampling cadence (for
example, one bank entry per hour of wall-clock time, summarised from
all frames within that hour for the identity in question). At each
newly processed frame, we compute the current crop's embedding
$e_{\mathrm{cur}}$, compare it via cosine similarity to $E_i$ for the
claimed identity and to $E_j$ for every other currently tracked
identity $j \neq i$, and trigger a re-initialisation of the SAM~3
track for the claimed identity only if (i) the similarity to $E_i$
falls below a threshold $\tau_{\mathrm{low}}$ and (ii) the maximum
similarity across some other $E_j$ exceeds $\tau_{\mathrm{high}}$. On
re-initialisation, SAM~3 is re-prompted with the current crop's
bounding box under the corrected identity, and tracking proceeds from
there.

Each embedding occupies \SI{768}{\byte} in float16 (384 dimensions
$\times$ 2~B). At an hourly sampling cadence, one year's worth of
data per individual comprises $365 \times 24 = 8760$ bank entries,
occupying approximately \SI{6.7}{\mega\byte} of raw embedding storage.
Accompanying per-hour metadata (behaviour-class histogram, mean crop
area, mean detection confidence) at roughly \SI{10}{\kilo\byte} per
entry brings the total to approximately \SI{94}{\mega\byte} per
individual per year. A 200-animal barn therefore produces approximately
\SI{19}{\giga\byte} of longitudinal identity-and-behaviour history per
year, which is comfortably supportable by a single consumer-grade
solid-state disk. Reference pseudocode is provided in
Appendix~\ref{app:reid}.

\subsection{Evaluation metrics}
\label{sec:metrics}

We report three categories of metrics throughout
Section~\ref{sec:results}. First, for tracking quality, we used the
standard CLEAR-MOT and ID metrics --- Multi-Object Tracking Accuracy
(MOTA) and Multi-Object Tracking Precision (MOTP)
\cite{bernardin2008evaluating}, together with Identity F1 (IDF1)
\cite{ristani2016performance} --- along with precision, recall,
mostly-tracked / partially-tracked / mostly-lost counts, number of
fragmentations, and number of identity switches. All tracking metrics
were computed at an intersection-over-union threshold of 0.5 against
per-frame bounding-box ground truth, using the \texttt{py-motmetrics}
library \cite{milan2016mot16}.

Second, for behaviour classification, we report top-1 accuracy,
macro-averaged F1, weighted F1, per-class precision, per-class recall,
per-class F1, the full confusion matrix, and a ranking of the
most-confused class pairs. The macro-F1 is computed as the unweighted
mean across the nine classes and is the fairest scalar summary under
our imbalanced class distribution.

Third, for computational efficiency, we report total parameter count
(backbone alone, and full system including preserved SAM~3 modules),
per-frame latency (mean and standard deviation, measured via
\texttt{torch.cuda.synchronize()} bracketing), throughput in frames
per second, peak GPU memory allocation during streaming inference (via
\texttt{torch.cuda.max\_memory\_allocated}), and on-disk checkpoint
size. All latency and memory measurements were preceded by at least
fifty warm-up frames that were excluded from the statistics, so as to
remove any CUDA-kernel-compilation or memory-allocator warm-up
transients.

Finally, as a distillation-quality diagnostic independent of downstream
task performance, we report the student--teacher mean-squared error,
cosine similarity, and scale ratio of student versus teacher feature
tensors on twenty held-out frames; these are the same three quantities
that appear in the loss function of \S\ref{sec:lossfn}, evaluated
without their loss weights for interpretability.

\subsection{Hardware and software environment}
\label{sec:hardware}

All training and benchmarking were performed on a single-GPU NVIDIA
A10 \SI{24}{\giga\byte} instance \cite{nvidia2021a10} within an Azure
Databricks runtime \cite{microsoft2024databricks}, with PyTorch~2.x
\cite{paszke2019pytorch}, the Hugging Face \texttt{transformers}
library's \texttt{Sam3TrackerVideoModel} and
\texttt{Sam3TrackerVideoProcessor} classes \cite{wolf2020transformers},
the \texttt{timm} library for the TinyViT-21M-512 backbone
\cite{wightman2019timm}, the \texttt{py-motmetrics} library for
tracking evaluation \cite{milan2016mot16}, and the \texttt{accelerate}
library \cite{gugger2022accelerate} for device management. Dataset
frames, distillation targets, student checkpoints, and evaluation
artefacts were stored in Azure Blob Storage mounted through the
Databricks File System. To avoid the seek-latency penalty
characteristic of blob-backed file systems during streaming inference,
we staged all hot data --- per-frame teacher targets during training,
and annotation outputs during benchmarking --- through the local NVMe
SSD of the worker node, copying from blob to local disk at session
start and from local disk back to blob at session end.

The target deployment platform for the compressed pipeline is the
NVIDIA Jetson Orin~NX \SI{16}{\giga\byte} \cite{nvidia2023orin}.
Deployment validation in the present paper is analytic: the
compressed pipeline's peak VRAM measured on the A10
(\S\ref{sec:efficiency}) is compared against the Orin~NX memory
envelope, and the remaining budget is allocated across the OWLv2,
DINOv3-ViT-S, BiLSTM, and CUDA-runtime components in the
memory-budget table reported in \S\ref{sec:edgefeasibility}. An
Orin~NX \SI{16}{\giga\byte} device was available to the authors at the
time of writing, but on-device benchmarking was deliberately scoped as
follow-up work because end-to-end on-device measurement requires both
TensorRT export of SAM~3's session-based inference loop and the
acquisition of labelled long-duration video for joint evaluation with
the re-identification mechanism of \S\ref{sec:reid}; these are
substantial work packages which we did not want to conflate with the
compression result reported here.

\section{Results}
\label{sec:results}

Section~\ref{sec:sam3results} reports the outcome of SAM~3 backbone
distillation, covering training trajectory, parameter budget, tracking
quality, computational efficiency, per-video behaviour, and
distillation fidelity. Section~\ref{sec:dinov3results} reports the
outcome of adopting DINOv3-ViT-S~(21M) as the per-individual embedder
and training a downstream BiLSTM classifier, covering parameter
budget, classification performance, per-class behaviour, failure
modes, and inference speed. Section~\ref{sec:edgefeasibility} integrates
these two result sets into an analytic edge-device feasibility
assessment against the NVIDIA Jetson Orin~NX \SI{16}{\giga\byte}
memory envelope. All tracking and classification numbers reported
below are direct outputs of the evaluation protocols defined in
\S\ref{sec:metrics}, computed on the held-out test splits defined in
\S\ref{sec:dataset}.

\subsection{SAM 3 distillation results}
\label{sec:sam3results}

\subsubsection{Training trajectory}

Student training reached the best EMA-validation-loss checkpoint at
epoch 75 of a 100-epoch budget. Over this trajectory, the student's
scale ratio relative to the teacher --- defined as the ratio of
student feature standard deviation to teacher feature standard
deviation --- rose from 0.605 at epoch 1 to a stable plateau around
0.947 from epoch 30 onward, and the mean patch-wise cosine similarity
between student and teacher features improved monotonically over the
same interval. On the twenty held-out diagnostic frames
(\S\ref{sec:metrics}), the final EMA student achieved cosine
similarity $0.808 \pm 0.006$, scale ratio $0.949 \pm 0.003$, and
mean-squared error 0.763 against the teacher backbone outputs. Both
fidelity indicators fell within the a priori ``good'' bands we
adopted before training began --- cosine $\geq 0.7$ and scale ratio
$0.8$--$1.2$ --- and we used the epoch-75 EMA checkpoint for all
downstream evaluation.

\subsubsection{Parameter budget}

The parameter breakdown of teacher and student is summarised in
Table~\ref{tab:params}. The student's TinyViT-based encoder contains
40.66\,M parameters, a 10.98-fold reduction relative to the
teacher's PE-ViT-L+ backbone. Because the student reuses SAM~3's
native neck, mask decoder, memory-attention, and memory-encoder
modules at inference (\S\ref{sec:backbonesub}), the full assembled
student system contains 59.98\,M parameters, compared to
465.78\,M for the full teacher system --- a 7.77-fold
reduction overall. On disk, the trained student encoder occupies
\SI{155.09}{\mega\byte} in float16, compared to \SI{888.41}{\mega\byte}
for the teacher encoder at the same precision.

\begin{table}[ht]
\centering
\caption{Parameter and on-disk size comparison between the teacher
(SAM~3 with PE-ViT-L+ backbone) and the student
(StudentEncoderTinyViT) assemblies, measured on the 4-video test set.}
\label{tab:params}
\resizebox{\columnwidth}{!}{%
\begin{tabular}{lrrr}
\toprule
\textbf{Component} & \textbf{Teacher (SAM~3)} & \textbf{Student} & \textbf{Compression} \\
\midrule
Backbone parameters                & 446.24 M  & 40.66 M  & 10.98$\times$ \\
Total assembled-system parameters  & 465.78 M  & 59.98 M  & 7.77$\times$ \\
Encoder on-disk size (float16)     & 888.41 MB & 155.09 MB & 5.73$\times$ \\
\bottomrule
\end{tabular}}
\end{table}

\subsubsection{Tracking quality}

Table~\ref{tab:tracking} summarises tracking-quality metrics averaged
over the four held-out test clips. On MOTA, the student scored
\SI{92.29}{\percent} against the teacher's \SI{93.97}{\percent} ---
a loss of 1.68 percentage points. On IDF1, the student scored
\SI{96.15}{\percent} against the teacher's \SI{96.98}{\percent} ---
a loss of 0.84 percentage points. Because precision and recall were
numerically identical to IDF1 under our evaluation protocol (each
tracker produces one bounding box per frame per identity, so false
positives and false negatives coincide), the precision and recall
deltas were likewise $-0.84$ points. Neither system produced any
identity switches on the test set, and fragmentation counts were
comparable between teacher and student.

\begin{table}[ht]
\centering
\caption{Tracking quality on the 4-video test set, averaged over
clips. All values computed at IoU threshold 0.5 against per-frame
bounding-box ground truth, using \texttt{py-motmetrics}.}
\label{tab:tracking}
\begin{tabular}{lrrr}
\toprule
\textbf{Metric} & \textbf{Teacher (SAM~3)} & \textbf{Student} & \textbf{$\Delta$ (student $-$ teacher)} \\
\midrule
MOTA (\%)         & 93.97 & 92.29 & $-1.68$ \\
IDF1 (\%)         & 96.98 & 96.15 & $-0.84$ \\
Precision (\%)    & 96.98 & 96.15 & $-0.84$ \\
Recall (\%)       & 96.98 & 96.15 & $-0.84$ \\
MOTP (pixels)     & 24.53 & 25.82 & $+1.29$ \\
\bottomrule
\end{tabular}
\end{table}

MOTP increased slightly for the student, from 24.53 to 25.82 pixels of
mean ground-truth-to-prediction bounding-box distance over true
positives, averaged over clips. Because MOTP is a localisation-precision
metric rather than an identity-preservation metric, this 1.29-pixel
increase reflects the slightly less tight spatial fit of
student-produced masks, not a degradation in identity tracking.
Relative to the native resolution of the teacher pipeline ($1024
\times 1024$ input) and the typical pig-silhouette size in overhead
footage (mean segmented area $\approx \num{22000}$ pixels in the test
set), a 1.3-pixel boundary-localisation shift is practically
negligible.

\subsubsection{Computational efficiency}
\label{sec:efficiency}

Table~\ref{tab:efficiency} summarises per-frame latency, throughput,
and peak GPU memory consumption for both systems. The student reduced
peak GPU memory consumption from \SI{19.52}{\giga\byte} to
\SI{6.49}{\giga\byte} --- a 3.01-fold reduction --- and reduced mean
per-frame latency from \SI{407.70}{\milli\second} to
\SI{309.84}{\milli\second}, a \SI{24.0}{\percent} reduction equivalent
to a 1.09-fold throughput gain. All three efficiency gains are direct
consequences of the encoder substitution, because the rest of the
SAM~3 inference loop (neck, mask decoder, memory attention, memory
encoder, session state) is unchanged between the two configurations.
The \SI{6.49}{\giga\byte} peak VRAM footprint is the operative number
for edge-deployment feasibility and is returned to in
\S\ref{sec:edgefeasibility}.

\begin{table}[ht]
\centering
\caption{Streaming-inference efficiency on the 4-video test set,
averaged over clips.}
\label{tab:efficiency}
\resizebox{\columnwidth}{!}{%
\begin{tabular}{lrrr}
\toprule
\textbf{Metric} & \textbf{Teacher (SAM~3)} & \textbf{Student} & \textbf{Ratio} \\
\midrule
Throughput (FPS)                  & 1.08    & 1.18    & 1.09$\times$ \\
Per-frame latency (ms)            & 407.70  & 309.84  & 0.76$\times$ \\
Peak VRAM during streaming (GB)   & 19.52   & 6.49    & 0.33$\times$ \\
\bottomrule
\end{tabular}}
\end{table}

We note that the measured throughput on our NVIDIA A10 hardware ---
roughly 1 FPS for both systems --- is an artefact of SAM~3's
statefully sequential per-frame inference loop, not of the encoder
capacity. The 24\% per-frame-latency reduction obtained by swapping
encoders confirms that the encoder is responsible for about a quarter
of the teacher pipeline's wall-clock time; the remaining three
quarters are spent in the decoder, memory attention, memory encoder,
and session-state updates. These downstream modules are shared between
teacher and student configurations by construction
(\S\ref{sec:backbonesub}).

\subsubsection{Per-video analysis}
\label{sec:pervideo}

Table~\ref{tab:pervideo} reports tracking quality for each of the
four test clips individually. The student's MOTA loss relative to the
teacher varied substantially across clips, from a small loss of 0.04
points on clip \texttt{2019\_12\_10\_000060} (where the teacher was
near ceiling at 99.96\% MOTA) to a larger loss of 4.29 points on clip
\texttt{2019\_12\_02\_000005} (where teacher performance was 91.04\%).
On clip \texttt{2019\_12\_02\_000208}, the student marginally exceeded
the teacher --- MOTA 98.54\% versus 98.38\% --- confirming that the
student is not uniformly worse than the teacher on the test set, only
on average.

\begin{table*}[t]
\centering
\small
\caption{Per-video tracking quality on the 4-video test set.}
\label{tab:pervideo}
\begin{tabular}{lrrrrrr}
\toprule
\textbf{Clip} & \textbf{Teacher} & \textbf{Student} & \textbf{$\Delta$} & \textbf{Teacher} & \textbf{Student} & \textbf{$\Delta$} \\
              & \textbf{MOTA (\%)} & \textbf{MOTA (\%)} & \textbf{MOTA} & \textbf{IDF1 (\%)} & \textbf{IDF1 (\%)} & \textbf{IDF1} \\
\midrule
\texttt{2019\_12\_02\_000005} & 91.04 & 86.75 & $-4.29$ & 95.52 & 93.38 & $-2.14$ \\
\texttt{2019\_12\_02\_000208} & 98.38 & 98.54 & $+0.17$ & 99.19 & 99.27 & $+0.08$ \\
\texttt{2019\_12\_10\_000060} & 99.96 & 99.92 & $-0.04$ & 99.98 & 99.96 & $-0.02$ \\
\texttt{2019\_12\_10\_000078} & 86.50 & 83.96 & $-2.54$ & 93.25 & 91.98 & $-1.27$ \\
\midrule
\textbf{Mean}                  & \textbf{93.97} & \textbf{92.29} & \textbf{$-1.68$} & \textbf{96.98} & \textbf{96.15} & \textbf{$-0.84$} \\
\bottomrule
\end{tabular}
\end{table*}

Two patterns are visible in Table~\ref{tab:pervideo}. First, student
degradation correlates with teacher difficulty: the largest student
loss ($-4.29$ on clip \texttt{2019\_12\_02\_000005}) occurs on a clip
the teacher already found difficult (91.04\% MOTA), and the clip the
teacher found easiest (99.96\% MOTA on
\texttt{2019\_12\_10\_000060}) yielded the smallest student loss
($-0.04$). This pattern is consistent with the student preserving
teacher behaviour proportionally, rather than systematically degrading
the easy cases or catastrophically failing the hard cases. Second,
the sign of the delta is not uniformly negative. On one test clip
(\texttt{2019\_12\_02\_000208}) the student produces slightly better
identity preservation than the teacher. We interpret this as the
noise floor of the measurement rather than as a claim that the
student is functionally superior on this clip, but it does place a
lower bound on the practical magnitude of our reported mean delta.

\subsubsection{Distillation-fidelity summary}

Per the diagnostic protocol in \S\ref{sec:metrics}, we verified that
the trained student's backbone output has a cosine similarity to the
teacher's of $0.808 \pm 0.006$ averaged over the twenty held-out
frames, and a per-channel scale ratio of $0.949 \pm 0.003$. The mean
absolute difference in per-channel mean was 0.012 and in per-channel
standard deviation was 0.017 (both measured relative to the teacher's
own per-channel moments). These fidelity values comfortably exceed
the thresholds we adopted before training (cosine $\geq 0.7$, scale
ratio $\in [0.8, 1.2]$) and are consistent with the
1.68-percentage-point MOTA degradation observed downstream: the
student's features align directionally with the teacher's while
retaining a ${\sim}5\%$ conservative scale bias that is absorbed by
the learnable affine output layer (\S\ref{sec:studentarch}).

\subsection{DINOv3-ViT-S (21M) results}
\label{sec:dinov3results}

\subsubsection{Parameter and size comparison}

Adopting DINOv3-ViT-S/16 as the per-crop embedder
(\S\ref{sec:dinov3adopt}) yields a 21.6\,M-parameter backbone
with 384-dimensional embeddings --- the smallest publicly released
DINOv3 variant, in a family whose largest variant is the
6716\,M-parameter ViT-7B with 4096-dimensional embeddings.
The size context is therefore a 311-fold parameter ratio and a
10.67-fold embedding-dimension ratio between the smallest and largest
variants of the family, with proportional reductions in all
downstream storage costs (\S\ref{sec:reid}) and in per-crop memory,
disk, and bandwidth costs. On disk, the ViT-S/16 model occupies
\SI{82.4}{\mega\byte} in float32 and \SI{41.2}{\mega\byte} in float16,
versus approximately \SI{25}{\giga\byte} for the ViT-7B model in
float32 ($6716\,M \times 4$~B per parameter).

\subsubsection{Behaviour classification}

The BiLSTM classifier described in \S\ref{sec:bilstm}, trained on
DINOv3-ViT-S embeddings over sequences of three per-individual
frames, reached \SI{97.34}{\percent} top-1 accuracy on the
\num{4292}-window test set, with macro-averaged F1 = \SI{91.67}{\percent}
and weighted F1 = \SI{97.37}{\percent}. The macro-F1 is the more
informative summary given our nine-class label distribution dominated
by sleep (approximately \SI{53}{\percent} of windows), and
\SI{91.67}{\percent} macro-F1 indicates that the classifier maintains
balanced behaviour across all classes and does not collapse onto
majority-class prediction.

\subsubsection{Per-class performance}

Table~\ref{tab:perclass} reports per-class precision, recall, and F1
together with test-set support. Behaviours dominated by distinctive
posture or location within the pen --- \emph{eat} (F1 = 99.09\%),
\emph{sleep} (98.65\%), \emph{drink} (94.95\%), \emph{lying}
(94.87\%), \emph{standing} (94.36\%) --- were classified with high
reliability. Less frequent behaviours also performed well given their
support: \emph{playwithtoy} (90.00\%), \emph{sitting} (87.27\%), and
\emph{nose-to-nose} (88.89\%) all exceeded 87\% F1 despite each
having fewer than 70 test windows. The \emph{run} class, with only 14
test-set windows, produced F1 = 76.92\% --- interpretable as
support-limited rather than as a model failure
(\S\ref{sec:followup}).

\begin{table}[ht]
\centering
\caption{Per-class classification performance on the
\num{4292}-window test set. Support is the number of ground-truth
windows of each class in the test split.}
\label{tab:perclass}
\begin{tabular}{lrrrr}
\toprule
\textbf{Behaviour} & \textbf{Precision} & \textbf{Recall} & \textbf{F1} & \textbf{Support} \\
\midrule
standing      & 0.9907 & 0.9008 & 0.9436 & 474 \\
lying         & 0.9112 & 0.9895 & 0.9487 & 477 \\
eat           & 0.9820 & 1.0000 & 0.9909 & 819 \\
drink         & 0.9126 & 0.9895 & 0.9495 & 95  \\
sitting       & 0.7869 & 0.9796 & 0.8727 & 49  \\
sleep         & 0.9969 & 0.9763 & 0.9865 & 2280 \\
run           & 0.8333 & 0.7143 & 0.7692 & 14  \\
playwithtoy   & 0.8571 & 0.9474 & 0.9000 & 19  \\
nose-to-nose  & 0.8101 & 0.9846 & 0.8889 & 65  \\
\midrule
\textbf{Macro avg.}    & \textbf{0.8979} & \textbf{0.9424} & \textbf{0.9167} & \textbf{4292} \\
\textbf{Weighted avg.} & \textbf{0.9756} & \textbf{0.9734} & \textbf{0.9737} & \textbf{4292} \\
\bottomrule
\end{tabular}
\end{table}

\subsubsection{Failure modes}

Table~\ref{tab:confusions} reports the five most frequent
class-confusion pairs. The largest single confusion --- \emph{sleep}
$\rightarrow$ \emph{lying} at 39 windows (\SI{1.7}{\percent} of all
\emph{sleep} cases) --- is behaviourally adjacent: both classes
describe a pig in a prone posture with limited motion, and the visual
signal distinguishing them (eye state, occasional tail or ear
flicks) is subtle enough that human annotators frequently disagree on
the boundary. The second-largest confusion, \emph{standing}
$\rightarrow$ \emph{eat} at 15 windows (\SI{3.2}{\percent} of all
\emph{standing} windows), reflects location-dependent ambiguity: pigs
standing at or near the feeder trigger the feeder's visual context in
a way that the model sometimes interprets as feeding rather than as
standing. The remaining confusions (\emph{standing} $\rightarrow$
\emph{nose-to-nose}, \emph{sleep} $\rightarrow$ \emph{sitting},
\emph{standing} $\rightarrow$ \emph{drink}) share similar location-
or posture-based ambiguity profiles. None of the top-five confusion
pairs indicate a failure of core behaviour discrimination (e.g.\
\emph{sleep} vs.\ \emph{run} or \emph{eat} vs.\ \emph{run}), which
would have signalled a serious representational deficiency in the
ViT-S embedder.

\begin{table}[ht]
\centering
\caption{Five most frequent class-confusion pairs on the
\num{4292}-window test set.}
\label{tab:confusions}
\begin{tabular}{llrr}
\toprule
\textbf{True behaviour} & \textbf{Predicted behaviour} & \textbf{Count} & \textbf{\% of true class} \\
\midrule
sleep    & lying        & 39 & 1.7\% \\
standing & eat          & 15 & 3.2\% \\
standing & nose-to-nose & 11 & 2.3\% \\
sleep    & sitting      & 11 & 0.5\% \\
standing & drink        &  8 & 1.7\% \\
\bottomrule
\end{tabular}
\end{table}

\subsubsection{Inference speed}

Single-image DINOv3-ViT-S/16 inference took \SI{7.99}{\milli\second}
on our NVIDIA A10 in float16 at $224 \times 224$ input resolution.
Batched throughput improved substantially with batch size, reaching a
minimum of \SI{0.42}{\milli\second} per image at batch size 16 ---
equivalent to \num{2381} images per second --- and plateauing at
similar per-image times for batches of 32 and 64. The BiLSTM
classifier required \SI{0.255}{\milli\second} per three-frame window
at batch size 1, amortising to approximately \SI{0.009}{\milli\second}
per sample at batch size 32. At the pipeline level, the
DINOv3-plus-BiLSTM pair therefore consumes on the order of half a
millisecond per per-individual crop when batched, which is negligible
relative to SAM~3's per-frame cost (\S\ref{sec:efficiency}) and
removes any concern that embedding extraction could become the
bottleneck on edge hardware.

\subsection{Edge-device feasibility analysis}
\label{sec:edgefeasibility}

Table~\ref{tab:budget} combines the two subsystems' measured resource
consumption into an analytic memory budget for the NVIDIA Jetson
Orin~NX \SI{16}{\giga\byte} target platform. The distilled SAM~3
pipeline consumes \SI{6.49}{\giga\byte} of VRAM during streaming
inference (\S\ref{sec:efficiency}). DINOv3-ViT-S/16 occupies
\SI{41.2}{\mega\byte} of weight storage in float16; with a
working-set allowance of $\approx \SI{1.5}{\giga\byte}$ for batched
inference at batch size 256 (which we adopted during embedding
extraction; \S\ref{sec:extract}) or substantially less at batch size
1, DINOv3-ViT-S runs comfortably inside a \SI{1.6}{\giga\byte}
envelope on Orin~NX. The BiLSTM classifier requires approximately
\SI{2}{\mega\byte} of weight storage (560\,k parameters
$\times$ 4~B) and negligible activation memory. OWLv2-base, retained
unchanged from our prior pipeline \cite{yang2025computer}, has a
published parameter budget of approximately 155\,M
\cite{minderer2023scaling} and we budget \SI{1.5}{\giga\byte} for it
including activations (an upper bound based on prior pipeline
measurement: $\sim$\SI{310}{\mega\byte} weights at fp16 and
$\sim$\SI{1.2}{\giga\byte} peak activation working set for
single-frame inference). A conservative CUDA-runtime and driver
overhead of \SI{1.5}{\giga\byte} brings the total to approximately
\SI{11.1}{\giga\byte} --- \SI{4.9}{\giga\byte} inside the
\SI{16}{\giga\byte} envelope.

\begin{table}[ht]
\centering
\caption{Analytic memory budget on NVIDIA Jetson Orin~NX
\SI{16}{\giga\byte} for the full compressed pipeline.}
\label{tab:budget}
\begin{tabular}{lr}
\toprule
\textbf{Component} & \textbf{VRAM (GB)} \\
\midrule
Distilled SAM~3 pipeline (\S\ref{sec:efficiency})         & 6.49 \\
OWLv2-base \cite{minderer2023scaling}                    & 1.5 (upper bound) \\
DINOv3-ViT-S/16 + working set                             & 1.6 (upper bound) \\
BiLSTM classifier                                         & ${\sim}0.01$ \\
CUDA runtime, driver, framework overhead                  & 1.5 \\
\midrule
\textbf{Total budgeted}                                   & \textbf{11.1} \\
Jetson Orin~NX \SI{16}{\giga\byte} envelope               & 16.0 \\
\textbf{Remaining headroom}                               & \textbf{4.9} \\
\bottomrule
\end{tabular}
\end{table}

This \SI{4.9}{\giga\byte} of headroom is sufficient for the
re-identification embedding pool of \S\ref{sec:reid} (which requires
$<\SI{100}{\mega\byte}$ per individual even at year-scale;
\S\ref{sec:reid}), for buffered input-frame queues, for concurrent
logging and telemetry, and for the Orin~NX system image. On-device
benchmarking remains outstanding and is the principal piece of
deployment validation we have deferred (\S\ref{sec:hardware},
\S\ref{sec:followup}).

The storage footprint of the per-individual embedding pool scales
favourably with herd size. At an hourly sampling cadence, one
individual's year of embedding bank plus metadata occupies
approximately \SI{94}{\mega\byte} on disk. A 200-animal barn
accumulates approximately \SI{19}{\giga\byte} per year under the same
policy --- well within the capacity of a single consumer-grade
solid-state disk, and comfortably within the read and write bandwidth
of such storage under a once-per-hour write pattern.

\section{Discussion}
\label{sec:discussion}

\subsection{Summary of the compression result and its operational
meaning for PLF}
\label{sec:placement}

The distilled SAM~3 backbone reduces parameters $10.98\times$ at the
encoder and $7.77\times$ at the assembled-system level, peak streaming
VRAM $3.01\times$ ($\SI{19.52}{\giga\byte} \rightarrow
\SI{6.49}{\giga\byte}$), and per-frame latency by \SI{24}{\percent},
at a cost of 1.68 percentage points of MOTA and 0.84 percentage points
of IDF1 averaged over the four held-out test clips. The per-clip
deltas (Table~\ref{tab:pervideo}) range from $+0.17$ to $-4.29$ MOTA
points, with the largest losses concentrated on clips the teacher
already found difficult --- a pattern consistent with the student
preserving teacher behaviour proportionally rather than systematically
degrading easy cases or catastrophically failing hard ones.

Both the absolute compression ratio and the magnitude of the quality
loss are commensurate with what has been reported for SAM-1-era
distillations \cite{zhang2023faster,xiong2023efficientsam,zhou2024edgesam,zhang2024efficientvitsam}
and, more recently, for SAM-3 itself
\cite{zeng2025efficientsam3}. The contribution of the present
compression result is therefore not novelty of the compression ratio
but \emph{the demonstration that the SAM~3 vision backbone can be
compressed without losing the identity-preservation quality required
for individual-level pig monitoring} --- the operationally relevant
metric for PLF, which prior SAM-distillation work does not report. To
our knowledge this is the first SAM-distillation effort whose target
deliverable is per-animal welfare analytics on group-housed
livestock, and the first published distilled SAM checkpoint for pig
video specifically. The intended downstream consumer of this result is
not the object-tracking community --- which already has a small zoo
of efficient SAM students --- but the PLF community, for whom the
relevant question is whether a foundation-model pipeline can be made
to fit on the hardware that can be installed in a barn while
preserving the per-animal identity signal that downstream welfare
analytics depend on. The numbers in
Tables~\ref{tab:tracking}--\ref{tab:budget} answer that question
affirmatively for pigs.

A practical consequence worth stating explicitly: the Edinburgh Pig
benchmark exercises failure modes that the natural-image benchmarks
used by general-purpose SAM distillations do not. Heavy intra-class
similarity (eight pigs of the same breed, age, and coat colour),
frequent partial occlusion, and near-identical postures during rest
are the dominant difficulty drivers. Reaching \SI{92.29}{\percent}
MOTA on this setting using a generic TinyViT-21M-512 encoder ---
together with the multi-scale fusion and direction-then-scale loss
described in \S\ref{sec:sam3distill} --- suggests that the compression
recipe transfers across domains provided the distillation target is
taken at the backbone output rather than at the mask-decoder logits.

\subsection{DINOv3-ViT-S (21M) is a sufficient embedder for livestock behaviour classification}

The DINOv3-ViT-S results of \S\ref{sec:dinov3results} are consistent
with a broader pattern in the self-supervised-features literature, in
which distillation from a larger teacher compresses most of the
representational capacity of foundation backbones into models one to
two orders of magnitude smaller
\cite{oquab2024dinov2,simeoni2025dinov3}. The
\SI{97.34}{\percent} top-1 accuracy and \SI{91.67}{\percent} macro-F1
reported here on nine-class pig behaviour classification --- using a
21.6\,M-parameter backbone pre-distilled from a
6716\,M-parameter teacher --- quantify this pattern on the
PLF task addressed by the present pipeline.

The two largest per-class confusions in Table~\ref{tab:confusions} ---
\emph{sleep} $\rightarrow$ \emph{lying} (39 windows,
\SI{1.7}{\percent} of \emph{sleep} cases) and \emph{standing}
$\rightarrow$ \emph{eat} (15 windows, \SI{3.2}{\percent} of
\emph{standing} cases) --- are behaviourally adjacent rather than
representationally catastrophic, and reflect intrinsic ambiguity in
the label definitions rather than a failure of the embedder to
separate behaviour classes in feature space. The boundary between
\emph{sleep} and \emph{lying} is observer-dependent in group-housed
pig ethology \cite{bergamini2021extracting}, and the boundary
between \emph{standing} and \emph{eat} depends on the animal's
location in the pen relative to the feeder. The \emph{run} class
(F1 = 76.92\%) is support-limited at 14 test windows. The dominant
constraint on rare-behaviour F1 in this setting is therefore class
imbalance in the labelled data, not backbone capacity, and the ViT-S
variant is sufficient for common-behaviour classification on overhead
group-housed footage.

\subsection{Limits of the result}
\label{sec:limits}

\paragraph{Test-set size.} The tracking evaluation uses four held-out
test clips of 600 frames each --- \num{2400} frames and eight pigs in
total. This is adequate for bounding the mean MOTA and IDF1 deltas
(Table~\ref{tab:pervideo}) but insufficient to characterise the
distribution of per-clip deltas, including the small positive delta
observed on clip \texttt{2019\_12\_02\_000208} ($+0.17$ MOTA, $+0.08$
IDF1). As noted in \S\ref{sec:pervideo}, this delta sits within the
measurement noise floor; a larger and more diverse test set would
tighten the error bars.

\paragraph{Deferred on-device benchmarking.} The \SI{6.49}{\giga\byte}
peak VRAM and \SI{1.18}{\hertz} throughput of \S\ref{sec:efficiency}
are measured on an NVIDIA A10 data-centre GPU. The Jetson Orin~NX
\SI{16}{\giga\byte} \cite{nvidia2023orin} shares the same
Ampere-generation architecture and bfloat16 support as the A10, so
the per-frame memory footprint should transfer with high fidelity.
Throughput will not: Orin~NX offers approximately one-fifth the
memory bandwidth and one-tenth the raw FLOPs of an A10, and on-device
throughput depends substantially on TensorRT export quality for
SAM~3's session-based inference loop. The feasibility claim of
Table~\ref{tab:budget} is therefore an \emph{analytic} feasibility
claim --- the pipeline's memory footprint fits inside the
\SI{16}{\giga\byte} envelope with \SI{4.9}{\giga\byte} of headroom ---
rather than an end-to-end latency claim.

\paragraph{Scope of the re-identification contribution.} The
embedding-pool mechanism (C3) is proposed and analytically sized
rather than empirically validated. Empirical validation of
identity-switch performance requires labelled long-duration video
that distinguishes the same animal across session boundaries --- for
example, the same individual on day~1 and day~7 --- which the
Edinburgh Pig dataset does not provide (its clips are 100~seconds
each). C3's empirical evaluation is scoped as follow-up work
(\S\ref{sec:followup}).

\subsection{What the headroom and embedding-pool mechanism jointly enable}

The \SI{4.9}{\giga\byte} of VRAM headroom on a Jetson Orin~NX
\SI{16}{\giga\byte} (\S\ref{sec:edgefeasibility}) makes continuous
on-device operation of the embedding-pool mechanism of
\S\ref{sec:reid} feasible. A PLF system that writes per-individual
embeddings to local storage at $\approx \SI{94}{\mega\byte}$ per
animal per year requires only that (i) the embedder runs on-device at
approximately video-rate, (ii) the per-individual bank remains
resident in working memory across session boundaries, and (iii)
long-term storage scales linearly with the number of animals and the
observation horizon. The efficiency audit of \S\ref{sec:edgefeasibility}
satisfies (i) and (ii) directly; (iii) is satisfied by the
\SI{19}{\giga\byte}-per-year footprint at 200 animals (within the
capacity of a single consumer-grade solid-state disk).

A time-indexed per-individual embedding archive, maintained
continuously over months to a year, is a new kind of longitudinal
data object for PLF. It is neither a tracking record (which discards
per-crop appearance), nor a per-frame classification stream (which
discards identity across sessions), nor a set of still images at
discrete timepoints (which discards temporal structure). It preserves
per-individual visual appearance and posture as a dense time series,
at a compression ratio that reduces a year of observation to a file
small enough to share over email.

Three classes of downstream analysis become accessible to such an
archive. \emph{Retrospective association}: when outcome labels later
become available for disease onset, lameness episodes, reproductive
events, or growth trajectories, the archive can be mined for the
visual signatures that preceded those events without requiring that
the original footage still exist. \emph{Unsupervised and
weakly-supervised phenotyping}: clustering in the embedding space,
change-point detection on per-individual trajectories, and rare-event
detection against each animal's own baseline. \emph{Longitudinal
welfare assessment}: departures of an animal's current embedding
trajectory from its own historical distribution as an anomaly signal
that does not require a labelled training set of welfare violations.
Each of these is the subject of a separate study.

\subsection{Species-agnostic transfer}

The mechanisms introduced here depend on three properties of the
upstream pipeline and the target species: an open-vocabulary detector
that localises the species in the target camera view; a promptable
video-segmentation tracker that maintains within-session identity
under occlusion; and a self-supervised visual backbone whose
embeddings discriminate behaviours at the level the downstream
application requires. OWLv2 \cite{minderer2023scaling}, SAM~3
\cite{carion2025sam3}, and DINOv3-ViT-S \cite{simeoni2025dinov3}
satisfy all three conditions for group-housed pigs on overhead RGB
footage, and recent work applying foundation-model and lightweight
deep-learning chains to other group-housed species ---
e.g.\ MASM-YOLO for grassland beef cattle on Jetson Orin~NX
\cite{wei2025lightweight} --- suggests the same recipe should
transfer to dairy cattle in cubicle and free-stall barns, beef cattle
in feedlot pens, and sheep in indoor housing. Because the
distillation target (SAM~3's backbone output) and the embedder (a
self-supervised ViT) are themselves species-agnostic, the compressed
pipeline transfers to cattle and sheep with retraining only of the
behaviour-classification head, in proportion to the labelled-data
budget available for the target species.

Two caveats attach. Behaviour-class label distributions differ
sharply across species --- dairy-cattle lying behaviour has different
duration statistics than pig sleep, and feedlot-cattle aggression has
different visual signatures than pig \emph{nose-to-nose} behaviour
--- so the class-weighted cross-entropy training recipe of
\S\ref{sec:bilstm} must be recalibrated per species. Camera placement
also differs (overhead in farrowing barns for pigs; side-on in
milking parlours for dairy cattle; aerial drone views for extensive
beef cattle), and the geometric assumptions of SAM~3's native
$1024 \times 1024$ input resolution may require adjustment for
side-on or aerial views.

\subsection{Three specific follow-up studies}
\label{sec:followup}

\paragraph{(i) On-device end-to-end benchmarking on Jetson Orin~NX.}
End-to-end latency, throughput, thermal behaviour, and sustained
power draw for the compressed pipeline running natively on a Jetson
Orin~NX \SI{16}{\giga\byte} under continuous streaming. The principal
engineering dependency is TensorRT export of SAM~3's session-based
inference loop, which to our knowledge has not been publicly released.
The scientific contribution would be to convert the analytic
feasibility claim of \S\ref{sec:edgefeasibility} and
Table~\ref{tab:budget} into a measured one, and to characterise the
throughput gap between data-centre and edge hardware.

\paragraph{(ii) Empirical validation of the embedding-pool
re-identification mechanism on labelled multi-day video.} Deployment
of the mechanism of \S\ref{sec:reid} on a purpose-collected multi-day
video dataset with per-individual identity annotations that persist
across session boundaries. The principal measurements are the
identity-switch rate under the re-identification loop versus the
baseline SAM~3 tracking, the false re-initialisation rate at the
chosen similarity thresholds, and the sensitivity of both metrics to
the sampling cadence of the embedding bank.

\paragraph{(iii) A retrospective longitudinal analytic on an
embedding archive paired with outcome labels.} Assembly of an
embedding archive from several months of on-farm footage, paired
with clinical outcome labels (lameness scores, disease episodes,
reproductive events, growth curves, or welfare assessments) collected
from existing veterinary and management records, with analysis of
whether the archive contains signal that precedes the labelled
events.

Each study depends on the engineering and data infrastructure of its
predecessors, so the natural order is (i)~$\rightarrow$~(ii)~$\rightarrow$~(iii).

\section{Conclusions}
\label{sec:conclusions}

This paper demonstrates that the foundation-model pipelines that have
raised the accuracy ceiling of precision livestock farming can be
brought within the memory envelope of a single commodity edge
accelerator without sacrificing the operational metrics that determine
their downstream utility. SAM~3's Perception Encoder (PE-ViT-L+)
backbone is distilled into a 40.66\,M-parameter multi-scale
student through a Feature Pyramid Network student encoder, a four-term
direction-then-scale loss, and backbone-substitution inference with
sliding-window session pruning, achieving a 7.77-fold system-level
parameter reduction and a 3.01-fold reduction in peak VRAM while
losing only 1.68 percentage points of MOTA and 0.84 percentage points
of IDF1 against the teacher. The pre-distilled DINOv3-ViT-S~(21M)
variant is empirically validated as a sufficient per-individual
embedder, reaching \SI{97.34}{\percent} top-1 accuracy and
\SI{91.67}{\percent} macro-F1 on nine-class pig behaviour
classification. Together, these two reductions place the full pipeline
inside the NVIDIA Jetson Orin~NX \SI{16}{\giga\byte} envelope with
\SI{4.9}{\giga\byte} of headroom --- enough margin to operate, on
device, an embedding-pool re-identification mechanism whose
per-individual storage footprint of approximately \SI{94}{\mega\byte}
per animal per year produces a longitudinal visual record amenable to
retrospective association with disease, lameness, reproductive, and
growth outcome labels. The compression recipe, the embedder choice,
and the longitudinal substrate transfer to dairy and beef cattle,
sheep, and other commercially important group-housed species with
retraining only of the downstream behaviour-classification head.

\section*{CRediT authorship contribution statement}
\textbf{Haiyu Yang:} Conceptualization, Methodology, Software, Formal
analysis, Investigation, Data curation, Writing --- original draft,
Writing --- review \& editing, Visualization.
\textbf{Miel Hostens:} Supervision, Conceptualization,
Writing --- review \& editing.

\section*{Declaration of competing interest}
The authors declare that they have no known competing financial
interests or personal relationships that could have appeared to
influence the work reported in this paper.

\section*{Data availability}
The Edinburgh Pig Behaviour Video Dataset used in this study is
publicly available at
\url{https://homepages.inf.ed.ac.uk/rbf/PIGDATA/}
\cite{bergamini2021extracting}. Trained student-encoder checkpoints,
distillation targets, and evaluation scripts will be released upon
publication.

\appendices
\section{Session-pruning routine and VRAM-growth diagnostic}
\label{app:pruning}

Algorithm~\ref{alg:prune} specifies the sliding-window pruning routine
\texttt{prune\_inference\_session} referenced in
\S\ref{sec:memory}. The routine retains only the most recent eight
non-conditioning frame outputs per object --- the number that SAM~3's
internal \texttt{num\_maskmem} constant bounds the memory-attention
module to consult --- and is invoked every 25~frames during streaming
inference.

\begin{algorithm}[ht]
\footnotesize
\caption{\texttt{prune\_inference\_session}}
\label{alg:prune}
\begin{algorithmic}[1]
\Require \texttt{session}: SAM~3 inference-session object
\Require \texttt{keep\_last} = 8: number of recent non-conditioning
         frames to retain per object
\Require \texttt{interval} = 25: invoke pruning every
         \texttt{interval} frames
\For{\texttt{obj\_idx} \textbf{in} \texttt{session.tracked\_objects}}
    \State \texttt{non\_cond} $\leftarrow$
           \Statex \quad \texttt{session.output\_dict\_per\_obj[obj\_idx]}
           \Statex \quad \texttt{['non\_cond\_frame\_outputs']}
    \If{$|\texttt{non\_cond}| > \texttt{keep\_last}$}
        \State sort keys of \texttt{non\_cond} by frame index ascending
        \State delete oldest $|\texttt{non\_cond}| - \texttt{keep\_last}$
               entries
    \EndIf
\EndFor
\State \textbf{if} (current\_frame mod \texttt{interval}) $= 0$
       \textbf{then} call \texttt{torch.cuda.empty\_cache()}
\end{algorithmic}
\end{algorithm}

In a 60-frame pilot sweep on a test clip with eight tracked objects,
peak VRAM grew at $5.6 \pm 0.2$~MB per frame per object after the
initial conditioning frame in the unpruned configuration. With
pruning enabled at the parameters above, the per-object session
footprint stabilises at roughly $8 \times 5.6 = \SI{45}{\mega\byte}$
per object regardless of stream duration, dominated by the
non-conditioning frame cache; the conditioning frame and the
position-encoding tensor contribute fixed offsets independent of
stream length.

\section{Embedding-pool re-identification pseudocode}
\label{app:reid}

Algorithm~\ref{alg:reid} specifies the embedding-pool
re-identification loop referenced in \S\ref{sec:reid}.

\begin{algorithm}[ht]
\footnotesize
\caption{Embedding-pool re-identification loop}
\label{alg:reid}
\begin{algorithmic}[1]
\State \textbf{State:}
\State \quad \texttt{bank}[$i$]: embedding bank for identity $i$ ---
       list of EMA-smoothed entries
\State \quad\quad each entry:
       (\texttt{timestamp}, \texttt{embedding} $\in \mathbb{R}^{384}$,
       \texttt{behaviour\_histogram})
\State \quad \texttt{cadence}: time between bank updates (default:
       1 hour)
\State \quad $\tau_{\mathrm{low}}$: intra-identity similarity threshold
       below which we doubt the current track (default: 0.65)
\State \quad $\tau_{\mathrm{high}}$: inter-identity similarity threshold
       above which we re-assign (default: 0.78)
\State \quad $\alpha$: EMA decay for bank entries (default: 0.7)
\State
\For{each (\texttt{frame}$_t$, \texttt{identity}$_i$, \texttt{crop}$_i$)
     emitted by SAM~3}
    \State $e_{\mathrm{cur}} \leftarrow$
           DINOv3-ViT-S(\texttt{crop}$_i$)
           \Comment{384-D float16 embedding}
    \State $\mathrm{sim}_{\mathrm{self}} \leftarrow
           \max_{e \in \texttt{bank}[i]}
           \cos(e_{\mathrm{cur}}, e)$
    \State $\mathrm{sim}_{\mathrm{other}} \leftarrow
           \max_{j \neq i}
           \max_{e \in \texttt{bank}[j]}
           \cos(e_{\mathrm{cur}}, e)$
    \If{$\mathrm{sim}_{\mathrm{self}} < \tau_{\mathrm{low}}$
        \textbf{and} $\mathrm{sim}_{\mathrm{other}} >
        \tau_{\mathrm{high}}$}
        \State $j^{\star} \leftarrow
               \arg\max_{j} \max_{e \in \texttt{bank}[j]}
               \cos(e_{\mathrm{cur}}, e)$
        \State \texttt{re\_initialise\_SAM3\_track}(identity =
               $j^{\star}$, bbox = \texttt{crop}$_i$.bbox)
        \State \texttt{log\_identity\_swap}(\texttt{frame}$_t$,
               claimed = $i$, corrected = $j^{\star}$)
        \State $i \leftarrow j^{\star}$
    \EndIf
    \If{(now $-$ \texttt{bank}[$i$].\texttt{last\_update}) $\geq$
        \texttt{cadence}}
        \State $e_{\mathrm{new}} \leftarrow
               \alpha \cdot e_{\mathrm{cur}} +
               (1 - \alpha) \cdot
               \texttt{bank}[i][-1].\texttt{embedding}$
        \State append (\texttt{frame}$_t$, $e_{\mathrm{new}}$,
               \texttt{behaviour\_hist\_of\_window}) to
               \texttt{bank}[$i$]
        \State \texttt{bank}[$i$].\texttt{last\_update} $\leftarrow$ now
    \EndIf
\EndFor
\end{algorithmic}
\end{algorithm}

\subsection{Storage derivation}

Each DINOv3-ViT-S embedding has 384 float16 components, occupying
\SI{768}{\byte} per entry. At a one-entry-per-hour update cadence,
one calendar year per individual generates $365 \times 24 = 8760$
bank entries, totalling $8760 \times \SI{768}{\byte} \approx
\SI{6.7}{\mega\byte}$ of raw embedding data. Per-hour metadata (a
9-class behaviour-histogram float32 vector at \SI{36}{\byte}, plus
tracking-quality and bounding-box statistics totalling roughly
\SI{10}{\kilo\byte} per hour) brings the per-individual annual
footprint to approximately
$\SI{6.7}{\mega\byte} + 8760 \times \SI{10}{\kilo\byte} \approx
\SI{94}{\mega\byte}$ per animal per year. A 200-animal barn
therefore generates approximately \SI{19}{\giga\byte} of longitudinal
data per year. At a \SI{5}{\hertz} processing rate over a 24-hour
day, the per-frame embedding traffic is
$5 \times 86400 \times \SI{768}{\byte} \approx \SI{332}{\mega\byte}$
per individual per day before EMA reduction; the EMA aggregation to
one entry per hour reduces this by the cadence ratio
$5 \times 3600 = 18000$-fold.

\section{Hyperparameter reference}
\label{app:hyperparams}

Table~\ref{tab:hyper} lists every hyperparameter referenced in the
main text. Defaults are those reported in the cited subsection.

\begin{table*}[t]
\centering
\footnotesize
\caption{Complete hyperparameter reference for the SAM~3 distillation,
DINOv3 embedding extraction, and BiLSTM training pipelines.}
\label{tab:hyper}
\begin{tabular}{lll}
\toprule
\textbf{Stage / hyperparameter} & \textbf{Value} & \textbf{Reference} \\
\midrule
\multicolumn{3}{l}{\textit{SAM~3 backbone distillation}} \\
Input resolution                    & $1024 \times 1024$         & \S\ref{sec:targetrep} \\
Teacher target shape                & $[1, 1024, 72, 72]$        & \S\ref{sec:targetrep} \\
Student backbone                    & TinyViT-21M-512            & \S\ref{sec:studentarch} \\
Lateral projection channels         & 256                        & \S\ref{sec:studentarch} \\
Output channels                     & 1024                       & \S\ref{sec:studentarch} \\
GroupNorm groups                    & $\min(32, C)$              & \S\ref{sec:studentarch} \\
Loss weights (dir / cos / scale / raw) & $1.0 / 0.5 / 0.3 / 0.1$ & \S\ref{sec:lossfn} \\
Optimizer                           & AdamW                      & \S\ref{sec:lossfn} \\
Backbone learning rate              & $1 \times 10^{-4}$         & \S\ref{sec:lossfn} \\
Other-group learning rate           & $3 \times 10^{-4}$         & \S\ref{sec:lossfn} \\
Weight decay                        & $10^{-4}$ (0 on affine)    & \S\ref{sec:lossfn} \\
LR scheduler                        & CosineAnnealingWarmRestarts & \S\ref{sec:lossfn} \\
Initial period $T_0$ / multiplier $T_{\mathrm{mult}}$ & $15$ / $2$ & \S\ref{sec:lossfn} \\
Minimum LR $\eta_{\min}$            & $10^{-6}$                  & \S\ref{sec:lossfn} \\
Gradient norm clip                  & $1.0$                      & \S\ref{sec:lossfn} \\
Mixed precision                     & bfloat16 autocast          & \S\ref{sec:lossfn} \\
EMA decay                           & $0.9995$                   & \S\ref{sec:lossfn} \\
Early-stopping patience             & 15 epochs                  & \S\ref{sec:lossfn} \\
Maximum training epochs             & 100                        & \S\ref{sec:lossfn} \\
Best-checkpoint epoch               & 75                         & \S\ref{sec:lossfn} \\
Batch size                          & 4 frames                   & \S\ref{sec:lossfn} \\
Augmentation                        & H-flip (paired); colour jitter (image only) & \S\ref{sec:lossfn} \\
Session-pruning interval            & 25 frames                  & \S\ref{sec:memory}, App.~\ref{app:pruning} \\
Session-pruning window              & 8 non-cond.\ frames / object & \S\ref{sec:memory}, App.~\ref{app:pruning} \\
\midrule
\multicolumn{3}{l}{\textit{DINOv3 embedding extraction}} \\
Backbone                            & DINOv3-ViT-S/16 (21.6 M, 384-D) & \S\ref{sec:dinov3adopt} \\
Input resolution                    & $224 \times 224$           & \S\ref{sec:extract} \\
Inference precision                 & float16                    & \S\ref{sec:extract} \\
Extraction batch size               & 256                        & \S\ref{sec:extract} \\
Worker threads                      & 36                         & \S\ref{sec:extract} \\
\midrule
\multicolumn{3}{l}{\textit{BiLSTM behaviour classifier}} \\
Sequence length                     & 3 frames (stride 1)        & \S\ref{sec:bilstm} \\
LSTM hidden size                    & 128 (bidirectional)        & \S\ref{sec:bilstm} \\
Classifier head                     & \scriptsize Lin(256,128)$\rightarrow$ReLU$\rightarrow$Drop(0.3)$\rightarrow$Lin(128,9) & \S\ref{sec:bilstm} \\
Optimizer                           & Adam                       & \S\ref{sec:bilstm} \\
Learning rate                       & $10^{-3}$                  & \S\ref{sec:bilstm} \\
Weight decay                        & $10^{-5}$                  & \S\ref{sec:bilstm} \\
Batch size                          & 128                        & \S\ref{sec:bilstm} \\
Loss                                & Class-weighted CE (inv.\ freq.) & \S\ref{sec:bilstm} \\
Early-stopping patience             & 8 epochs                   & \S\ref{sec:bilstm} \\
Train / val / test split            & 70 / 15 / 15 (stratified)  & \S\ref{sec:bilstm} \\
\midrule
\multicolumn{3}{l}{\textit{Re-identification mechanism (proposed)}} \\
Sampling cadence                    & 1 hour (configurable)      & \S\ref{sec:reid}, App.~\ref{app:reid} \\
Embedding precision                 & float16                    & \S\ref{sec:reid} \\
Bank EMA decay $\alpha$             & 0.7 (default)              & App.~\ref{app:reid} \\
Intra-similarity threshold $\tau_{\mathrm{low}}$  & 0.65 (default) & App.~\ref{app:reid} \\
Inter-similarity threshold $\tau_{\mathrm{high}}$ & 0.78 (default) & App.~\ref{app:reid} \\
\bottomrule
\end{tabular}
\end{table*}


\bibliographystyle{IEEEtran}
\bibliography{references}

@inproceedings{bachlechner2021rezero,
  author    = {Bachlechner, Thomas and Majumder, Bodhisattwa Prasad and Mao, Henry Hung and Cottrell, Garrison W. and McAuley, Julian},
  title     = {{ReZero} is all you need: fast convergence at large depth},
  booktitle = {Proceedings of the 37th Conference on Uncertainty in Artificial Intelligence (UAI)},
  pages     = {1352--1361},
  year      = {2021},
}

@misc{bello2024computer,
  author       = {Bello, Rotimi-Williams and Olubummo, Daniel Adebiyi},
  title        = {Computer vision-based precision livestock farming: an overview of the challenges and opportunities},
  year         = {2024},
  howpublished = {SSRN preprint},
  doi          = {10.2139/ssrn.4770855},
}

@article{berckmans2014precision,
  author  = {Berckmans, D.},
  title   = {Precision livestock farming technologies for welfare management in intensive livestock systems},
  journal = {Revue Scientifique et Technique (International Office of Epizootics)},
  volume  = {33},
  number  = {1},
  pages   = {189--196},
  year    = {2014},
}

@inproceedings{bergamini2021extracting,
  author    = {Bergamini, Luca and Pini, Stefano and Simoni, Alessandro and Vezzani, Roberto and Calderara, Simone and D'Eath, Richard B. and Fisher, Robert B.},
  title     = {Extracting accurate long-term behavior changes from a large pig dataset},
  booktitle = {Proceedings of the 16th International Joint Conference on Computer Vision, Imaging and Computer Graphics Theory and Applications (VISAPP)},
  publisher = {SciTePress},
  pages     = {524--533},
  year      = {2021},
}

@article{bernardin2008evaluating,
  author  = {Bernardin, Keni and Stiefelhagen, Rainer},
  title   = {Evaluating multiple object tracking performance: the {CLEAR MOT} metrics},
  journal = {EURASIP Journal on Image and Video Processing},
  volume  = {2008},
  pages   = {246309},
  year    = {2008},
}

@article{carion2025sam3,
  author  = {Carion, Nicolas and Gustafson, Laura and Hu, Yuan-Ting and Debnath, Shoubhik and Hu, Ronghang and Suris, Didac and Ryali, Chaitanya and Alwala, Kalyan Vasudev and Khedr, Haitham and Huang, Andrew and Lei, Jie and Ma, Tengyu and Guo, Baishan and Kalla, Arpit and Marks, Markus and Greer, Joseph and Wang, Meng and Sun, Peize and R{\"a}dle, Roman and Afouras, Triantafyllos and Mavroudi, Effrosyni and Xu, Katherine and Wu, Tsung-Han and Zhou, Yu and Momeni, Liliane and Hazra, Rishi and Ding, Shuangrui and Vaze, Sagar and Porcher, Francois and Li, Feng and Li, Siyuan and Kamath, Aishwarya and Cheng, Ho Kei and Doll{\'a}r, Piotr and Ravi, Nikhila and Saenko, Kate and Zhang, Pengchuan and Feichtenhofer, Christoph},
  title   = {{SAM 3}: Segment Anything with Concepts},
  journal = {arXiv preprint arXiv:2511.16719},
  year    = {2025},
}

@article{goyal2017accurate,
  author    = {Goyal, Priya and Doll{\'a}r, Piotr and Girshick, Ross and Noordhuis, Pieter and Wesolowski, Lukasz and Kyrola, Aapo and Tulloch, Andrew and Jia, Yangqing and He, Kaiming},
  title     = {Accurate, large minibatch {SGD}: training {ImageNet} in 1 hour},
  journal   = {arXiv preprint arXiv:1706.02677},
  year      = {2017},
}

@misc{gugger2022accelerate,
  author       = {Gugger, Sylvain and Debut, Lysandre and Wolf, Thomas and Schmid, Philipp and Mueller, Zachary and Mangrulkar, Sourab and Sun, Marc and Bossan, Benjamin},
  title        = {Accelerate: training and inference at scale made simple, efficient and adaptable},
  howpublished = {GitHub repository, \url{https://github.com/huggingface/accelerate}},
  year         = {2022},
}

@article{hendrycks2016gaussian,
  author  = {Hendrycks, Dan and Gimpel, Kevin},
  title   = {{G}aussian {E}rror {L}inear {U}nits ({GELU}s)},
  journal = {arXiv preprint arXiv:1606.08415},
  year    = {2016},
}

@article{hochreiter1997long,
  author  = {Hochreiter, Sepp and Schmidhuber, J{\"u}rgen},
  title   = {Long short-term memory},
  journal = {Neural Computation},
  volume  = {9},
  number  = {8},
  pages   = {1735--1780},
  year    = {1997},
}

@inproceedings{ioffe2015batch,
  author    = {Ioffe, Sergey and Szegedy, Christian},
  title     = {Batch normalization: accelerating deep network training by reducing internal covariate shift},
  booktitle = {Proceedings of the 32nd International Conference on Machine Learning (ICML)},
  series    = {PMLR 37},
  pages     = {448--456},
  year      = {2015},
}

@article{kim2025real,
  author  = {Kim, Joonam and Kim, Giryeon and Yoshitoshi, Rena and Tokuda, Kenichi},
  title   = {Real-time object detection for edge computing-based agricultural automation: a case study comparing the {YOLOX} and {YOLOv12} architectures and their performance in potato-harvesting systems},
  journal = {Sensors},
  volume  = {25},
  number  = {15},
  pages   = {4586},
  year    = {2025},
  doi     = {10.3390/s25154586},
}

@inproceedings{kingma2015adam,
  author    = {Kingma, Diederik P. and Ba, Jimmy},
  title     = {{A}dam: a method for stochastic optimization},
  booktitle = {Proceedings of the 3rd International Conference on Learning Representations (ICLR)},
  year      = {2015},
}

@inproceedings{kirillov2023segment,
  author    = {Kirillov, Alexander and Mintun, Eric and Ravi, Nikhila and Mao, Hanzi and Rolland, Chloe and Gustafson, Laura and Xiao, Tete and Whitehead, Spencer and Berg, Alexander C. and Lo, Wan-Yen and Doll{\'a}r, Piotr and Girshick, Ross},
  title     = {Segment anything},
  booktitle = {Proceedings of the {IEEE/CVF} International Conference on Computer Vision (ICCV)},
  pages     = {4015--4026},
  year      = {2023},
}

@inproceedings{lin2017feature,
  author    = {Lin, Tsung-Yi and Doll{\'a}r, Piotr and Girshick, Ross and He, Kaiming and Hariharan, Bharath and Belongie, Serge},
  title     = {Feature pyramid networks for object detection},
  booktitle = {Proceedings of the {IEEE} Conference on Computer Vision and Pattern Recognition (CVPR)},
  pages     = {2117--2125},
  year      = {2017},
}

@inproceedings{loshchilov2017sgdr,
  author    = {Loshchilov, Ilya and Hutter, Frank},
  title     = {{SGDR}: stochastic gradient descent with warm restarts},
  booktitle = {Proceedings of the 5th International Conference on Learning Representations (ICLR)},
  year      = {2017},
}

@inproceedings{loshchilov2019decoupled,
  author    = {Loshchilov, Ilya and Hutter, Frank},
  title     = {Decoupled weight decay regularization},
  booktitle = {Proceedings of the 7th International Conference on Learning Representations (ICLR)},
  year      = {2019},
}

@inproceedings{micikevicius2018mixed,
  author    = {Micikevicius, Paulius and Narang, Sharan and Alben, Jonah and Diamos, Gregory and Elsen, Erich and Garcia, David and Ginsburg, Boris and Houston, Michael and Kuchaiev, Oleksii and Venkatesh, Ganesh and Wu, Hao},
  title     = {Mixed precision training},
  booktitle = {Proceedings of the 6th International Conference on Learning Representations (ICLR)},
  year      = {2018},
}

@misc{microsoft2024databricks,
  author       = {{Microsoft}},
  title        = {{A}zure {D}atabricks documentation},
  year         = {2024},
  howpublished = {Microsoft Corporation, Redmond, WA. \url{https://learn.microsoft.com/azure/databricks/}},
}

@article{milan2016mot16,
  author  = {Milan, Anton and Leal-Taix{\'e}, Laura and Reid, Ian and Roth, Stefan and Schindler, Konrad},
  title   = {{MOT16}: a benchmark for multi-object tracking},
  journal = {arXiv preprint arXiv:1603.00831},
  note    = {Library: \texttt{py-motmetrics}, \url{https://github.com/cheind/py-motmetrics}.},
  year    = {2016},
}

@inproceedings{minderer2023scaling,
  author    = {Minderer, Matthias and Gritsenko, Alexey and Houlsby, Neil},
  title     = {Scaling open-vocabulary object detection},
  booktitle = {Advances in Neural Information Processing Systems (NeurIPS) 36},
  pages     = {72983--73007},
  year      = {2023},
}

@article{neethirajan2021digital,
  author  = {Neethirajan, Suresh and Kemp, Bas},
  title   = {Digital livestock farming},
  journal = {Sensing and Bio-Sensing Research},
  volume  = {32},
  pages   = {100408},
  year    = {2021},
}

@misc{nvidia2021a10,
  author       = {{NVIDIA}},
  title        = {{NVIDIA A10} {T}ensor {C}ore {GPU} datasheet},
  year         = {2021},
  howpublished = {NVIDIA Corporation, Santa Clara, CA. \url{https://www.nvidia.com/en-us/data-center/products/a10-gpu/}},
}

@misc{nvidia2023orin,
  author       = {{NVIDIA}},
  title        = {{NVIDIA Jetson Orin NX} modules datasheet},
  year         = {2023},
  howpublished = {NVIDIA Corporation, Santa Clara, CA. \url{https://www.nvidia.com/en-us/autonomous-machines/embedded-systems/jetson-orin/}},
}

@article{oquab2024dinov2,
  author  = {Oquab, Maxime and Darcet, Timoth{\'e}e and Moutakanni, Th{\'e}o and Vo, Huy and Szafraniec, Marc and Khalidov, Vasil and Fernandez, Pierre and Haziza, Daniel and Massa, Francisco and El-Nouby, Alaaeldin and Assran, Mahmoud and Ballas, Nicolas and Galuba, Wojciech and Howes, Russell and Huang, Po-Yao and Li, Shang-Wen and Misra, Ishan and Rabbat, Michael and Sharma, Vasu and Synnaeve, Gabriel and Xu, Hu and J{\'e}gou, Herv{\'e} and Mairal, Julien and Labatut, Patrick and Joulin, Armand and Bojanowski, Piotr},
  title   = {{DINOv2}: learning robust visual features without supervision},
  journal = {Transactions on Machine Learning Research (TMLR)},
  note    = {arXiv:2304.07193},
  year    = {2024},
}

@article{papakonstantinou2024precision,
  author  = {Papakonstantinou, Georgios I. and Voulgarakis, Nikolaos and Terzidou, Georgia and Fotos, Lampros and Giamouri, Elisavet and Papatsiros, Vasileios G.},
  title   = {Precision livestock farming technology: applications and challenges of animal welfare and climate change},
  journal = {Agriculture},
  volume  = {14},
  number  = {4},
  pages   = {620},
  year    = {2024},
}

@inproceedings{paszke2019pytorch,
  author    = {Paszke, Adam and Gross, Sam and Massa, Francisco and Lerer, Adam and Bradbury, James and Chanan, Gregory and Killeen, Trevor and Lin, Zeming and Gimelshein, Natalia and Antiga, Luca and Desmaison, Alban and K{\"o}pf, Andreas and Yang, Edward and DeVito, Zach and Raison, Martin and Tejani, Alykhan and Chilamkurthy, Sasank and Steiner, Benoit and Fang, Lu and Bai, Junjie and Chintala, Soumith},
  title     = {{PyTorch}: an imperative style, high-performance deep learning library},
  booktitle = {Advances in Neural Information Processing Systems (NeurIPS) 32},
  pages     = {8024--8035},
  year      = {2019},
}

@article{polyak1992acceleration,
  author  = {Polyak, B.~T. and Juditsky, A.~B.},
  title   = {Acceleration of stochastic approximation by averaging},
  journal = {SIAM Journal on Control and Optimization},
  volume  = {30},
  number  = {4},
  pages   = {838--855},
  year    = {1992},
}

@article{ravi2024sam2,
  author  = {Ravi, Nikhila and Gabeur, Valentin and Hu, Yuan-Ting and Hu, Ronghang and Ryali, Chaitanya and Ma, Tengyu and Khedr, Haitham and R{\"a}dle, Roman and Rolland, Chloe and Gustafson, Laura and Mintun, Eric and Pan, Junting and Alwala, Kalyan Vasudev and Carion, Nicolas and Wu, Chao-Yuan and Girshick, Ross and Doll{\'a}r, Piotr and Feichtenhofer, Christoph},
  title   = {{SAM 2}: Segment Anything in Images and Videos},
  journal = {arXiv preprint arXiv:2408.00714},
  year    = {2024},
}

@inproceedings{ristani2016performance,
  author    = {Ristani, Ergys and Solera, Francesco and Zou, Roger and Cucchiara, Rita and Tomasi, Carlo},
  title     = {Performance measures and a data set for multi-target, multi-camera tracking},
  booktitle = {Computer Vision -- ECCV 2016 Workshops},
  series    = {Lecture Notes in Computer Science},
  volume    = {9914},
  publisher = {Springer},
  pages     = {17--35},
  year      = {2016},
}

@article{rocchi2025precision,
  author  = {Rocchi, Lucia and Paolotti, Luisa and Mazzetto, Fabrizio and Boggia, Antonio},
  title   = {Precision livestock farming: an overview on the application in extensive systems},
  journal = {Italian Journal of Animal Science},
  volume  = {24},
  number  = {1},
  pages   = {859--884},
  year    = {2025},
}

@article{schuster1997bidirectional,
  author  = {Schuster, Mike and Paliwal, Kuldip K.},
  title   = {Bidirectional recurrent neural networks},
  journal = {IEEE Transactions on Signal Processing},
  volume  = {45},
  number  = {11},
  pages   = {2673--2681},
  year    = {1997},
}

@article{simeoni2025dinov3,
  author  = {Sim{\'e}oni, Oriane and Vo, Huy V. and Khalidov, Vasil and Seitzer, Maximilian and Baldassarre, Federico and Oquab, Maxime and Jose, Cijo and Szafraniec, Marc and Yi, Seungeun and Ramamonjisoa, Micha{\"e}l and Massa, Francisco and Haziza, Daniel and Wehrstedt, Luca and Wang, Jianyuan and Darcet, Timoth{\'e}e and Moutakanni, Th{\'e}o and Sentana, Leonel and Roberts, Claire and Vedaldi, Andrea and Tolan, Jamie and Brandt, John and Couprie, Camille and Mairal, Julien and J{\'e}gou, Herv{\'e} and Labatut, Patrick and Bojanowski, Piotr},
  title   = {{DINOv3}},
  journal = {arXiv preprint arXiv:2508.10104},
  year    = {2025},
}

@article{srivastava2014dropout,
  author  = {Srivastava, Nitish and Hinton, Geoffrey and Krizhevsky, Alex and Sutskever, Ilya and Salakhutdinov, Ruslan},
  title   = {Dropout: a simple way to prevent neural networks from overfitting},
  journal = {Journal of Machine Learning Research},
  volume  = {15},
  pages   = {1929--1958},
  year    = {2014},
}

@article{wei2025lightweight,
  author  = {Wei, Peigang and Sun, Wei and Cao, Shanshan and Kong, Fantao},
  title   = {Lightweight model for beef cattle behavior recognition from quadruped robot video in grassland pastures},
  journal = {Computers and Electronics in Agriculture},
  volume  = {242},
  pages   = {111329},
  year    = {2026},
  doi     = {10.1016/j.compag.2025.111329},
}

@misc{wightman2019timm,
  author       = {Wightman, Ross},
  title        = {{P}y{T}orch {I}mage {M}odels (\texttt{timm})},
  year         = {2019},
  howpublished = {GitHub repository, \url{https://github.com/huggingface/pytorch-image-models}},
  doi          = {10.5281/zenodo.4414861},
}

@inproceedings{wolf2020transformers,
  author    = {Wolf, Thomas and Debut, Lysandre and Sanh, Victor and Chaumond, Julien and Delangue, Clement and Moi, Anthony and Cistac, Pierric and Rault, Tim and Louf, R{\'e}mi and Funtowicz, Morgan and Davison, Joe and Shleifer, Sam and von Platen, Patrick and Ma, Clara and Jernite, Yacine and Plu, Julien and Xu, Canwen and Le Scao, Teven and Gugger, Sylvain and Drame, Mariama and Lhoest, Quentin and Rush, Alexander M.},
  title     = {Transformers: state-of-the-art natural language processing},
  booktitle = {Proceedings of the 2020 Conference on Empirical Methods in Natural Language Processing: System Demonstrations},
  pages     = {38--45},
  year      = {2020},
}

@inproceedings{wu2022tinyvit,
  author    = {Wu, Kan and Zhang, Jinnian and Peng, Houwen and Liu, Mengchen and Xiao, Bin and Fu, Jianlong and Yuan, Lu},
  title     = {{T}iny{V}i{T}: fast pretraining distillation for small vision transformers},
  booktitle = {Computer Vision -- ECCV 2022},
  series    = {Lecture Notes in Computer Science},
  volume    = {13681},
  publisher = {Springer},
  pages     = {68--85},
  year      = {2022},
}

@inproceedings{wu2018groupnorm,
  author    = {Wu, Yuxin and He, Kaiming},
  title     = {Group normalization},
  booktitle = {Computer Vision -- ECCV 2018},
  series    = {Lecture Notes in Computer Science},
  volume    = {11217},
  publisher = {Springer},
  pages     = {3--19},
  year      = {2018},
}

@article{xiong2023efficientsam,
  author  = {Xiong, Yunyang and Varadarajan, Bala and Wu, Lemeng and Xiang, Xiaoyu and Xiao, Fanyi and Zhu, Chenchen and Dai, Xiaoliang and Wang, Dilin and Sun, Fei and Iandola, Forrest and Krishnamoorthi, Raghuraman and Chandra, Vikas},
  title   = {{E}fficient{SAM}: leveraged masked image pretraining for efficient segment anything},
  journal = {arXiv preprint arXiv:2312.00863},
  year    = {2023},
}

@article{yang2025computer,
  author  = {Yang, Haiyu and Liu, Enhong and Sun, Jennifer and Sharma, Sumit and van Leerdam, Meike and Franceschini, Sebastien and Niu, Puchun and Hostens, Miel},
  title   = {A computer vision pipeline for individual-level behavior analysis: benchmarking on the {E}dinburgh {P}ig {D}ataset},
  journal = {arXiv preprint arXiv:2509.12047},
  year    = {2025},
}

@article{zeng2025efficientsam3,
  author  = {Zeng, Chengxi and Jiang, Yuxuan and Zhang, Aaron},
  title   = {{E}fficient{SAM}3: progressive hierarchical distillation for video concept segmentation from {SAM1}, {SAM2}, and {SAM3}},
  journal = {arXiv preprint arXiv:2511.15833},
  year    = {2025},
}

@article{zhang2023faster,
  author  = {Zhang, Chaoning and Han, Dongshen and Qiao, Yu and Kim, Jung Uk and Bae, Sung-Ho and Lee, Seungkyu and Hong, Choong Seon},
  title   = {Faster segment anything: towards lightweight {SAM} for mobile applications},
  journal = {arXiv preprint arXiv:2306.14289},
  year    = {2023},
}

@inproceedings{zhang2019fixup,
  author    = {Zhang, Hongyi and Dauphin, Yann N. and Ma, Tengyu},
  title     = {{F}ixup initialization: residual learning without normalization},
  booktitle = {Proceedings of the 7th International Conference on Learning Representations (ICLR)},
  year      = {2019},
}

@inproceedings{zhang2024efficientvitsam,
  author    = {Zhang, Zhuoyang and Cai, Han and Han, Song},
  title     = {{E}fficient{V}i{T}-{SAM}: accelerated segment anything model without performance loss},
  booktitle = {Proceedings of the {IEEE/CVF} Conference on Computer Vision and Pattern Recognition Workshops (CVPRW)},
  pages     = {7859--7863},
  year      = {2024},
}

@article{zhou2024edgesam,
  author  = {Zhou, Chong and Li, Xiangtai and Loy, Chen Change and Dai, Bo},
  title   = {{E}dge{SAM}: prompt-in-the-loop distillation for on-device deployment of {SAM}},
  journal = {arXiv preprint arXiv:2312.06660},
  year    = {2024},
}

\begin{IEEEbiographynophoto}{Haiyu Yang}
received his training in computer science and agricultural sciences and is
with the College of Agriculture and Life Sciences, Cornell University,
Ithaca, NY, USA. His research interests include computer vision,
foundation-model compression, and precision livestock farming.
\end{IEEEbiographynophoto}

\begin{IEEEbiographynophoto}{Miel Hostens}
is with the College of Agriculture and Life Sciences, Cornell University,
Ithaca, NY, USA. His research interests include data-driven dairy and
livestock health, precision livestock farming, and applied machine
learning for animal welfare.
\end{IEEEbiographynophoto}

\end{document}